\documentclass[sigconf, 9pt]{acmart}
\AtBeginDocument{%
  }

\copyrightyear{2027}
\acmYear{2027}
\setcopyright{cc}
\setcctype{by}
\acmConference[SenSys '27]{ACM/IEEE International Conference on Embedded Artificial Intelligence and Sensing Systems}{May 17--20, 2027}{Boulder, CO, USA}
\acmBooktitle{ACM/IEEE International Conference on Embedded Artificial Intelligence and Sensing Systems (SenSys '27), May 17--20, 2027, Boulder, CO, USA}
\acmDOI{10.1145/3862575.3862832}
\acmISBN{979-8-4007-3124-2/2027/05}

\usepackage{algorithm}
\usepackage{algorithmic}
\usepackage{xspace} % to use xspace in newcommand
\usepackage{subcaption} % to use subfigure
\usepackage{amsmath} % to use mathbf
\usepackage{booktabs} % to use toprul, midrule, bootomrule
\usepackage{multirow} % to use multirow
\usepackage{makecell,enumitem} % to use in table
\usepackage[table]{xcolor}
\newcommand{\ours}{{CoDA}\xspace}

\begin{document}

%%
%% The "title" command has an optional parameter,
%% allowing the author to define a "short title" to be used in page headers.
\title{CoDA: Frequency-Asymmetric Co-Design for Adaptive Quantized Edge CNNs}

%%
%% The "author" command and its associated commands are used to define
%% the authors and their affiliations.
%% Of note is the shared affiliation of the first two authors, and the
%% "authornote" and "authornotemark" commands
%% used to denote shared contribution to the research.

%\author{
    %Authors
%    Anonymous submissions
%}

 \author{Yoojin Kwon}
 \authornote{These authors contributed equally to this research.}
% \orcid{1234-5678-9012}
\affiliation{%
   \institution{Seoul National University}
   \city{Seoul}
   \country{Republic of Korea}
}
 \email{ideastraw@snu.ac.kr}

\author{Hongjun Suh}
 \authornotemark[2]
\affiliation{%
   \institution{Seoul National University}
   \city{Seoul}
   \country{Republic of Korea}
}
\email{hjsuh319@snu.ac.kr}

\author{Wooseok Lee}
 \authornotemark[2]
\affiliation{%
   \institution{Seoul National University}
   \city{Seoul}
   \country{Republic of Korea}
}
\email{andylws@snu.ac.kr}

\author{Taesik Gong}
 \authornote{Corresponding authors}
\affiliation{%
   \institution{UNIST}
   \city{Ulsan}
   \country{Republic of Korea}
}
\email{taesik.gong@unist.ac.kr}

\author{Songyi Han}
 \authornote{Now at Google DeepMind}
\affiliation{%
   \institution{Google}
   \city{Mountain View} 
   \country{USA}
}
\email{syhan@google.com}

\author{Hyung-Sin Kim}
 \authornotemark[3]
\affiliation{%
   \institution{Seoul National University}
   \city{Seoul}
   \country{Republic of Korea}
}
\email{hyungkim@snu.ac.kr}

%%

%%
%% By default, the full list of authors will be used in the page
%% headers. Often, this list is too long, and will overlap
%% other information printed in the page headers. This command allows
%% the author to define a more concise list
%% of authors' names for this purpose.

%%
%% The code below is generated by the tool at http://dl.acm.org/ccs.cfm.
%% Please copy and paste the code instead of the example below.
%%


%%
%% Keywords. The author(s) should pick words that accurately describe
%% the work being presented. Separate the keywords with commas.
%% A "teaser" image appears between the author and affiliation
%% information and the body of the document, and typically spans the
%% page.
% \begin{teaserfigure}
%   \includegraphics[width=\textwidth]{sampleteaser}
%   \caption{Seattle Mariners at Spring Training, 2010.}
%   \Description{Enjoying the baseball game from the third-base
%   seats. Ichiro Suzuki preparing to bat.}
%   \label{fig:teaser}
% \end{teaserfigure}

% \received{20 February 2007}
% \received[revised]{12 March 2009}
% \received[accepted]{5 June 2009}

%%
%% This command processes the author and affiliation and title
%% information and builds the first part of the formatted document.

%

%%%%% Sections
\begin{abstract}
   Edge vision models must be both compact and adaptive: they are compressed before deployment, yet operate in sensing environments that shift after deployment. Quantization-aware training enables efficient inference but usually produces a static source-domain model, while test-time adaptation improves robustness but often requires backpropagation, mutable full-precision weights, or extra memory. 
   We present \ours, a frequency-asymmetric co-design for quantized, batch-normalized edge convolutional neural networks (CNNs). During training, \ours performs QAT on low-frequency reconstructions, encouraging low-bit models to learn domain-stable visual structure. During deployment, \ours performs Frequency-Aware Batch Normalization, a forward-only adaptation method that updates only normalization statistics from unlabeled target inputs, using high-frequency target cues to capture current sensing conditions. \ours requires no source data, labels, backpropagation, optimizer state, or weight updates. 
   Across corruption, rendition, and capture-condition (lighting) shifts, bit-widths, QAT methods, and CNN backbones, \ours improves accuracy by up to 8.63 pp on CIFAR10-C and 6.99 pp on ImageNet-C over forward-only TTA baselines. Beyond accuracy gains, \ours improves the accuracy-efficiency tradeoff with INT8 execution of compute-heavy layers on edge platforms.
   % Beyond accuracy gains, \ours improves accuracy-efficiency tradeoff with INT8 execution for compute-heavy convolution and linear layers during adaptive quantized inference on edge platforms.
   The code is available at https://github.com/YoojinKwon/CoDA/.
\end{abstract}
\maketitle
%%%%%%%%% Introduction 
\vspace{-2ex}
\section{Introduction}\label{sec:introduction}
Edge vision systems are increasingly expected to sense, decide, and adapt locally. Cameras on robots, drones, wearables, traffic sensors, agricultural monitors, IoT cameras, and augmented-reality devices must process visual streams under tight limits on latency, memory, and connectivity~\cite{zeng2024adapt,boldo2024domain,imam2024domain,antonazzi2024r2snet}.
Recent mobile and sensing systems have made substantial progress on efficient edge AI through low-power onboard vision~\cite{lee2025hypercam}, memory-adaptive inference~\cite{li2024flexnn}, dynamic sparsity~\cite{liu2024dynaspa}, GPU scheduling~\cite{wang2025agile3d}, heterogeneous processor execution~\cite{jung2025aria}, quantized MCU deployment~\cite{liu2026instmeter}, and NPU-centric on-device inference~\cite{yin2026shadownpu, li2026veclut}. 
However, most of them optimize the \textit{execution} of a deployed model. They do not fully address a complementary problem: \textbf{what happens when the sensing environment changes after the model has already been compressed and shipped?}

This problem is unavoidable in deployed sensing systems. Lighting, weather, motion blur, sensor noise, viewpoints, and backgrounds can shift after deployment~\cite{imagenet-es,yoon2026egoxtreme,ss6d,kim2026imagenet}.  Recent mobile and sensing systems have therefore begun to treat adaptation as a first-class problem~\cite{fang2024adashadow,liu2025daf,ma2023cost,wen2023adaptivenet,zhao2025responsive}.
% ~\cite{wen2023adaptivenet, ma2023cost, liu2025daf,fang2024adashadow,zhao2025responsive}. 
%
Most directly, BoTTA~\cite{danilowski2026botta} benchmarks on-device test-time adaptation (TTA) under practical mobile and edge constraints, including limited target samples, limited category exposure, diverse shifts, overlapping shifts, and real-device resource metrics. These works make clear that adaptation is no longer only an algorithmic robustness problem; it is a deployment systems problem~\cite{choi2026emerging,choiorganized}.

Yet there remains a gap between adaptive inference and quantized edge deployment. Many high-performing TTA methods require backpropagation, optimizer state, mutable parameters, or full-precision computation
~{\cite{zhang2023adanpc,wang2020tent,niu2022efficient,gong2022note,niu2023towards,zhang2022memo}}. 
In contrast, many quantized edge deployments expose a stricter contract: no labels, no source data, no backpropagation, no optimizer state, no weight updates, and quantized execution for the compute-heavy convolution and linear operators. Under this contract, adaptation cannot be treated as online training. Instead, it must occur through a small amount of mutable inference state.

Quantization-aware training (QAT) attacks the efficiency side of this problem. It trains low-bit models that reduce model size, memory, and inference cost~\cite{jacob2018quantization,esser2020learned,zhang2018lq,li2023vit}. However, QAT usually treats deployment as \textit{static} inference: the model is optimized for source-domain accuracy and then frozen. 
TTA attacks the robustness side by adapting a model using unlabeled target  inputs~\cite{nado2020evaluating,schneider2020robustness,wang2020tent,niu2022efficient,wang2022continual,niu2023towards,su2024unraveling}. Forward-only TTA methods are more compatible with edge deployment because they avoid backpropagation and update normalization statistics or other lightweight inference-time variables~\cite{nado2020evaluating,schneider2020robustness,su2024unraveling,niu2024forward,ma2026pea}. But they are still typically \textit{post hoc}: a model is trained or quantized first, and adaptation is added afterward.

This paper argues that compressed training and deployment-time adaptation should be \textbf{co-designed}. If a deployed edge model can only adapt through lightweight forward-pass inference state, then the representation learned before deployment should be shaped around that adaptation mechanism. 
We therefore do not treat QAT and TTA as independent stages, but assign distinct visual information to different stages of the deployment pipeline.

Our key observation is a \textbf{frequency-asymmetric division of labor}. Low-frequency image components tend to preserve shape and slowly varying semantic structure, making them suitable for training a low-bit model with limited representational capacity. High-frequency components often encode texture, blur, noise, compression artifacts, and sensor-specific details. These details can be unreliable to memorize during source-domain quantized training, but useful to observe at deployment because they describe the current target environment. 
Prior work has used frequency information for robustness analysis, domain adaptation, domain generalization, curriculum learning, spectral augmentation, and frequency-domain training~\cite{yin2019fourier,wang2020high,yang2020fda,huang2021fsdr,chattopadhyay2023pasta,wang2023efficienttrain,xu2020learning}. 
We use frequency differently: not as another augmentation or preprocessing, but as a division of labor between what the quantized model learns before deployment and what remains mutable after deployment.

We introduce~\textbf{\ours}, a frequency-asymmetric co-design for quantized, batch-normalized edge convolutional neural networks (CNNs) under deployment-time domain shift. 
During source-time training, \ours performs QAT on low-frequency reconstructions of the training images. This encourages the compressed model to allocate its limited bit budget to domain-stable visual structure rather than source-specific high-frequency artifacts. 
During deployment, \ours performs Frequency-Aware Batch Normalization (FABN), a forward-only adaptation mechanism that updates only normalization statistics from unlabeled target inputs. At the first normalization layer, FABN separates activations into low- and high-frequency components. Low-frequency statistics are updated conservatively to preserve source-trained semantic structure, while high-frequency statistics are estimated from the current target batch to capture deployment-specific sensing conditions. Later normalization layers use lightweight statistic adaptation for efficiency.

The novelty of \ours is not quantization alone, frequency filtering alone, or BN-statistic adaptation alone. The novelty is the \textbf{source-time/deployment-time co-design}. \ours trains the low-bit model to rely on low-frequency structure before deployment and reserves high-frequency target information for backpropagation-free adaptation after deployment. 
This distinguishes \ours from 
(1) efficient inference systems that optimize execution of fixed models, 
(2) adaptive edge systems that rely on training or runtime model management, 
(3) conventional QAT methods that compress models without preparing them for target-domain adaptation, 
(4) conventional TTA methods that adapt models post hoc, and
(5) frequency-domain robustness methods that use frequency mainly for augmentation, curriculum, analysis, or domain transfer.
\ours is scoped to batch-normalized CNNs~\cite{howard2019mobilenetv3,tan2019efficientnet}, widely used lightweight architectures for resource-constrained vision pipelines. \ours turns BN statistics into a lightweight adaptation interface. Convolution and linear layers remain quantized and unchanged, while only inference statistics are updated during forward passes. %LayerNorm-based vision transformers expose different normalization state and token structure, and recent work on integer-only ViT inference, forward-only ViT adaptation, and backpropagation-free embedding alignment addresses a related but different design space~\cite{li2023vit,niu2024forward,ma2026pea}. 
%This paper focuses on quantized CNNs, including lightweight architectures that remain widely used in resource-constrained vision pipelines~\cite{howard2019mobilenetv3,tan2019efficientnet}.

%Our extensive experiments demonstrate \ours's effectiveness: (1) across various bit-widths
% , including extremely low 2 bits
%; (2) on multiple architectures, such as ResNet as well as lightweight MobileNet and EfficientNet; and (3) under both episodic and continual shifts. 
%
%Figure~\ref{fig:intro}(a) highlights that \ours improves ImageNet-C accuracy 
%\textbf{+4.53-6.99 percentage points (pp)} over the forward-only baseline, across all evaluated architectures and bit-widths, including even in the FP models. 
%Furthermore, since FABN's frequency split is applied only in the first BN layer, the computational overhead is modest. Figure~\ref{fig:intro}(b) shows that when running INT8-accelerated ResNet50 inference on a Jetson Orin Nano CPU (batch size 16), INT8-\ours takes only 5.86 s versus 4.68 s for INT8-NORM, remaining \textbf{3.1$\times$ faster} than FP-NORM (18.24 s), \textbf{3.6$\times$ faster} than FP-TENT (21.16 s), and \textbf{7.2$\times$ faster} than FP-SAR (42.11 s). Excluding quantization, FP-\ours still provides much faster inference than FP-SAR due to its forward-only nature, without sacrificing accuracy.

Across corruption, rendition, and capture-condition (lighting) shifts, window sizes, bit-widths, QAT methods, CNN backbones, lightweight architectures, continual shifts, and edge devices, \ours improves robustness while preserving deployability. 
On ImageNet-C, \ours improves accuracy by 4.53--6.99 percentage points (pp) over the forward-only baseline, and it also improves CIFAR10-C accuracy by up to 8.63 pp. 
On Jetson Orin Nano and Raspberry Pi 4B CPU with INT8 ResNet18/50 inference, \ours preserves the efficiency advantage of quantized forward-only execution while avoiding the latency cost of backpropagation-based adaptation. 
In particular, the system overhead of \ours such as peak memory, CPU utilization is largely confined to the first-layer BN and remains bounded under small window sizes (e.g. $k$=4) while maintaining its robustness. 
These results show that adaptive inference does not have to become online training: with a frequency-aware interface between offline QAT and online statistic adaptation, edge CNNs can remain compact, efficient, and responsive to deployment-time domain shift.

This paper makes four contributions. 
\begin{itemize}[leftmargin=*]
    \item We formulate adaptive quantized inference as an embedded vision systems problem under a strict deployment contract: no labels, no source data, no backpropagation, no optimizer state, no weight updates, and quantized execution for compute-heavy layers.
    \item We propose \ours, a frequency-asymmetric co-design that uses low-frequency source information for quantized representation learning and high-frequency target information for forward-only normalization-statistic adaptation.
    \item We implement FABN as a lightweight adaptation interface for quantized, batch-normalized CNNs, updating only inference statistics while keeping convolution and linear operators quantized and unchanged.
    % \item We implement FABN as a lightweight adaptation interface for quantized CNNs, updating only inference statistics while keeping convolution and linear operators quantized and unchanged.
    \item We evaluate \ours across robustness benchmarks, low-bit settings, CNN architectures, continual shifts, batch-size settings, and execution on multiple edge platforms (i.e. Jetson Orin Nano and Raspberry Pi 4B), showing that quantized edge CNNs can improve robustness while preserving a fixed-weight, forward-only deployment path.
\end{itemize}
We next formalize this deployment contract and then show how \ours satisfies it by separating source-time quantized representation learning from deployment-time statistic adaptation.
\begin{table*}[t]
\centering
\caption{Deployment compatibility under the adaptive quantized inference contract. ``System-dependent'' indicates that compatibility depends on the runtime, backend, or implementation.}
\vspace{-2ex}
\label{tab:deployment_contract}
\resizebox{\linewidth}{!}{
\begin{tabular}{lccccccc}
\toprule
\textbf{Method family}
& \makecell{\textbf{Quant.} \\ \textbf{conv/linear}}
& \makecell{\textbf{No} \\ \textbf{backprop}}
& \makecell{\textbf{No} \\ \textbf{weight update}}
& \makecell{\textbf{No} \\ \textbf{optimizer}}
& \makecell{\textbf{No} \\ \textbf{source data}}
& \makecell{\textbf{Online} \\ \textbf{adaptation}}
& \makecell{\textbf{Co-designed} \\ \textbf{QAT+TTA}} \\
\midrule
Static QAT~\cite{jacob2018quantization,esser2020learned,zhang2018lq,jin2020neural}
& Yes & Yes & Yes & Yes & Yes & No & No \\ 
\midrule
\makecell[l]{Efficient edge inference systems \\
~\cite{lee2025hypercam,li2024flexnn,liu2024dynaspa,han2024pantheon,jung2025aria,sun2025adaptqnet,yin2026shadownpu}
}
& System-dependent & Yes & Yes & Yes & Yes & Often no & No \\
\makecell[l]{Adaptive edge systems \\
~\cite{fang2024adashadow, liu2025daf, ma2023cost, wen2023adaptivenet,zhao2025responsive}
}
% ~\cite{wen2023adaptivenet, ma2023cost, liu2025daf, fang2024adashadow, zhao2025responsive}}
& System-dependent & Often no & Often no & Often no & Usually yes & Yes & No \\
\midrule
\makecell[l]{Backprop-based TTA 
~\cite{wang2020tent,niu2022efficient,wang2022continual,niu2023towards,gong2022note,gong2023sotta}}
& Often no & No & No & No & Yes & Yes & No \\
\midrule
\makecell[l]{Forward-only BN adaptation
~\cite{nado2020evaluating,schneider2020robustness,su2024unraveling}
}
& Yes & Yes & Yes & Yes & Yes & Yes & No \\
\makecell[l]{Forward-only prompt/embedding \\ adaptation
~\cite{niu2024forward,ma2026pea}
}
& Model-dependent & Yes & Usually yes & Method-dependent & Yes & Yes & No \\
Quantized TTA~\cite{deng2025test}
& Yes/partial & Method-dependent & Method-dependent & Method-dependent & Yes & Yes & Often no \\
%On-device TTA benchmarks, e.g., BoTTA
%& Evaluation framework & Evaluation framework & Evaluation framework & Evaluation framework & Evaluation framework & Evaluation framework & Evaluation framework \\
\midrule
\textbf{CoDA}
& \textbf{Yes} & \textbf{Yes} & \textbf{Yes} & \textbf{Yes} & \textbf{Yes} & \textbf{Yes} & \textbf{Yes} \\
\bottomrule
\end{tabular}
}
\vspace{-2ex}
\end{table*}

\section{Deployment Model and Problem Statement}\label{sec:deploy}

Adaptive quantized inference is a deployment problem where edge vision models must remain \textit{efficient} after compression, yet \textit{adapt} when the environment changes. This section defines the deployment contract targeted by \ours and clarifies why existing efficient-inference and adaptation approaches do not jointly satisfy it.

%%%%%%%%%%%%%%%%%%%%%%%%%%%%%%%%%
%%%%%%%%%%%%%%%%%%%%%%%%%%%%%%%%%
\vspace{-1ex}
\subsection{Target Deployment Contract}

We consider embedded and mobile vision deployments in which a model is trained offline, compressed or quantized, and then deployed to a resource-constrained device that processes an unlabeled sensor stream. After deployment, the target distribution may shift because of lighting, weather, motion blur, sensor noise, compression artifacts or site-specific backgrounds~\cite{baek2026position,suh2026resource,imagenet-es}. The target stream may be approximately stationary over a short period, as in a fixed camera installed at a new site, or non-stationary, as in a robot, drone, wearable camera, or vehicle-mounted sensor.

We focus on an inference-oriented deployment contract:
\begin{itemize}[leftmargin=*]
    \item No target labels are available during deployment
    \item No raw source data are stored on the device
    \item No backpropagation is performed during deployment
    \item No optimizer state is maintained during deployment
    \item No model weights are updated during deployment
    \item Compute-heavy convolution and linear layers remain quantized
    \item Only lightweight inference state may be mutable
\end{itemize}
The deployed model artifact may include trained weights, quantization parameters, and source normalization statistics, but it does not retain raw source images or labels. This contract is \textit{stricter} than general on-device learning ~\cite{wang2022melon,lin2022device,shin2024effective}. It reflects deployments where the runtime is optimized for inference, gradient support is expensive or unavailable, source data cannot be stored for privacy or memory reasons, and model weights are expected to remain fixed after release.

Recent on-device TTA benchmarks such as BoTTA~\cite{danilowski2026botta} show that realistic edge adaptation should account for limited target samples, limited category exposure, diverse shifts, overlapping shifts, and real-device resource costs. We therefore model deployment using a \textbf{small unlabeled target-domain adaptation window}
\begin{equation}
W_{\tau} = \{x_{\tau-k+1}, \ldots, x_{\tau}\},   
\end{equation}
where $\tau$ is the deployment step.
$W_{\tau}$ may contain a single frame, a small streaming batch, or a periodically collected set of target samples. 
\ours updates only lightweight inference statistics from $W_{\tau}$, without labels, source data, gradients, optimizer state, or weight updates.

%%%%%%%%%%%%%%%%%%%%%%%%%%%%%%%%%%%%%%%%%%%
%%%%%%%%%%%%%%%%%%%%%%%%%%%%%%%%%%%%%%%%%%%
\subsection{Adaptive Quantized Inference}

Let $\mathcal{D}^{s} = \{(x_i^s, y_i^s)\}_{i=1}^{N}$
denote the labeled source data available before deployment, and let $\mathcal{X}^{t} = \{x_1^t, x_2^t, \ldots\}$ denote the unlabeled target stream observed after deployment. The source and target distributions may \textit{differ}, and the target distribution may also \textit{change} over time.
Offline training produces a quantized model $f_{\theta_q}$, where $\theta_q$ denotes fixed quantized weights and quantization parameters. During deployment, $\theta_q$ is not updated. The only mutable component is a small inference-time state variable $s_{\tau}$, which in \ours consists of normalization statistics and bounded statistic buffers.

At deployment step $\tau$, the system receives an unlabeled adaptation window $W_{\tau}$, updates the mutable inference state $s_{\tau}$ using only forward computation and predicts $\hat{y}_{\tau}^{t}$ for the current test data $x_{\tau}^{t}$:
\begin{equation}
s_{\tau} = U(s_{{\tau}-1}, W_{{\tau}}), \quad \hat{y}_{{\tau}}^{{t}} = f_{\theta_q}(x_{{\tau}}^{{t}}; s_{{\tau}})
\end{equation}
The update rule \(U\) must satisfy the deployment contract: it uses no labels, no source data, no gradients, no optimizer state, and no weight updates. The objective is to improve target-stream accuracy {under the device budgets, where predictions are generated using the inference states updated solely by \(U\) in Eq. (2)}:
{\begin{equation}
\max_{U} \ \mathrm{Acc}(f_{\theta_q}(\mathcal{X}^t))
\quad
\text{s.t.}
\quad
L \leq L_{\max}, \quad
M \leq M_{\max}, 
\end{equation}}
where \(L\) and \(M\) denote deployment-time latency and peak memory. This formulation makes adaptive quantized inference a systems problem: the best method is not simply the one with the highest unconstrained accuracy, but the one that improves robustness while preserving the deployability of quantized inference.

\begin{figure*}[t]
    \centering
    \includegraphics[width=0.9\linewidth]{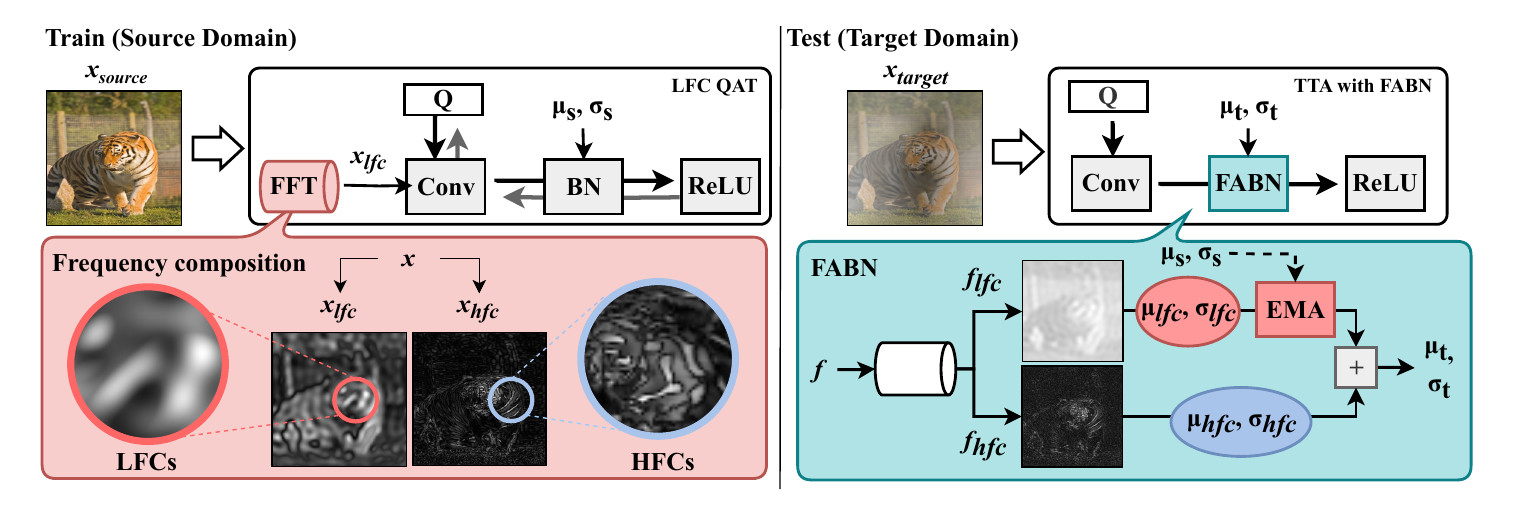} 
    \vspace{-5ex}
    \caption{
    An illustration of \ours. Using 2D Fourier transformation, an image is decomposed into HFCs, capturing fast-changing patterns (i.e., edges or stripes) and LFCs, capturing slow-changing patterns (i.e., smooth shape). During training (Left; LFC QAT), \ours learns generalizable features in LFCs rather than irregular patterns in HFCs. At test time under domain shift, \ours uses the full target activation to adapt the BN layer (Right; FABN); For LFC activations, FABN resets BN statistics to running estimates and updates gradually. For HFCs, FABN preserves the target distribution without aligning to source statistics. 
    }
    \Description{Two-panel system diagram. Left, training on the source domain: an input image passes through an FFT (Fast Fourier transformation) block that keeps its low-frequency components, which feed a quantized convolution, batch normalization, and ReLU pipeline trained with quantization-aware training. A zoomed inset shows the image split into a blurry low-frequency image and an edge-like high-frequency image. Right, test-time adaptation on the target domain: the same quantized pipeline, with batch normalization replaced by FABN (Frequency-aware Batch Normalization). Inside FABN, the feature map is split into low- and high-frequency parts; low-frequency statistics are updated through an exponential moving average, high-frequency statistics are taken from the current target batch, and both are combined into the normalization statistics.}
    \label{fig:method_overview}
    \vspace{-2.5ex}
\end{figure*}

%%%%%%%%%%%%%%%%%%%%%%%%%%%%%%%
%%%%%%%%%%%%%%%%%%%%%%%%%%%%%%%
\vspace{-1ex}
\subsection{Deployment Compatibility of Existing Approaches}

Table~\ref{tab:deployment_contract} summarizes representative method families under the adaptive quantized inference contract. Existing approaches solve important parts of the problem, but they do not jointly provide quantized execution, inference-only adaptation, fixed weights, source-free operation, online adaptation, and co-designed training/adaptation.

Static QAT~\cite{jacob2018quantization,esser2020learned,zhang2018lq,jin2020neural} provides efficient low-bit inference, but treats deployment as fixed-model execution and does not adapt to target-domain shifts. Efficient edge inference systems~\cite{lee2025hypercam,li2024flexnn,liu2024dynaspa,han2024pantheon,jung2025aria,sun2025adaptqnet,yin2026shadownpu} optimize latency, memory,  scheduling, or hardware mapping, but typically assume that the deployed model itself remains fixed. Adaptive edge-AI systems~[\citenum{fang2024adashadow},  {\citenum{liu2025daf}, \citenum{ma2023cost}}, \citenum{wen2023adaptivenet}, \citenum{zhao2025responsive}] 
% ~\cite{wen2023adaptivenet,ma2023cost, liu2025daf, fang2024adashadow, zhao2025responsive} 
respond to changing data or runtime conditions, but often require architecture search, online training, weight updates, expert selection, or cloud collaboration. Backpropagation-based TTA methods~\cite{wang2020tent,niu2022efficient,wang2022continual,niu2023towards,gong2022note,gong2023sotta} adapt to target inputs, but  gradients, optimizer state, mutable parameters, and full-precision computation conflict with static quantized inference. Forward-only TTA methods~\cite{nado2020evaluating,schneider2020robustness,su2024unraveling,niu2024forward,ma2026pea,deng2025test} are more deployment-compatible, but they are typically post hoc: a model is trained or quantized first, and adaptation is added afterward.

\ours targets the missing combination: a quantized CNN whose source-time representation is trained around the adaptation mechanism available after deployment, namely lightweight forward-pass inference statistics.

%%%%%%%%%%%%%%%%%%%%%%%%%%%%%%%%%%
%%%%%%%%%%%%%%%%%%%%%%%%%%%%%%%%%%
\vspace{-1.5ex}

\subsection{Design Requirements for \ours}
\label{subsec:design_requirements}

The deployment contract leads to three design requirements:
\begin{itemize}[leftmargin=*]
    \item Adaptation must occur through a small amount of mutable inference state rather than through model weights. %\ours therefore uses batch-normalization statistics as the adaptation interface. 

    \item The compute-heavy path, such as convolution and linear layers, must remain quantized. %\ours keeps convolution and linear layers quantized and unchanged during deployment. Frequency-aware processing is restricted to the first normalization layer, while later layers use lightweight statistic (BN) adaptation.

    \item Source-time training must prepare the model for deployment-time adaptation. A low-bit model has limited representational capacity, so it should not spend that capacity equally on all source-domain details. %\ours trains the quantized model on low-frequency source reconstructions to emphasize domain-stable structure, while reserving high-frequency target information for forward-only statistic adaptation after deployment.
\end{itemize}
%
% \vspace{-1ex}
These requirements motivate the \ours design described next: low-frequency QAT before deployment and frequency-aware batch-normalization adaptation after deployment.

%%%%%%%%%% Method
\vspace{-1ex}
\section{\ours: Co-Designed Quantized Training and Forward-Only Adaptation}\label{sec:method}

Guided by the deployment contract in Section~\ref{sec:deploy}, \ours separates what is \textit{learned before deployment} from what remains \textit{mutable after deployment}. \ours has two stages.
(1) Before deployment, \ours performs \textbf{low-frequency quantization-aware training (LFC-QAT)}, which trains a low-bit CNN on low-frequency reconstructions of labeled source images. This encourages the quantized representation to focus on \textit{domain-stable} visual structure. 
(2) During deployment, \ours keeps the quantized weights fixed and performs \textbf{Frequency-Aware Batch Normalization (FABN)}, a forward-only adaptation mechanism that updates only normalization statistics from unlabeled target windows. 
Figure~\ref{fig:method_overview} summarizes this source-time/deployment-time division of labor.

The design follows directly from the requirements in Section~\ref{sec:deploy}. 
%
% First, because deployment cannot update weights, adaptation must occur through mutable inference state; \ours uses BN statistics as this state. 
{First, because model weights remain fixed after deployment, CoDA uses BN statistics as lightweight mutable inference state.}
Second, because compute-heavy operators must remain quantized, \ours leaves convolution and linear layers unchanged during deployment. 
%
% Third, because target labels and source data are unavailable, source-time training must prepare the quantized model for the limited statistic-based adaptation available after deployment;
% \ours trains the quantized model on low-frequency source reconstructions to emphasize domain-stable structure.
{Third, source-time training prepares the model for adaptation through this restricted interface: LFC-QAT encourages the fixed quantized representation to rely on domain-stable low-frequency structure, allowing FABN to account for deployment-time sensing changes through activation statistics.}

\vspace{-2ex}
%%%%%%%%%%%%%%%%%%%%%%%%%%%%%%%
%%%%%%%%%%%%%%%%%%%%%%%%%%%%%%%
\subsection{Frequency Interface}

\ours uses frequency decomposition as the interface between offline quantized training and online statistic adaptation. \ours applies a 2D Fourier transformation \(\mathcal{F}(\cdot)\) on an input image \(x\), converting its spatial pixel patterns into multiple frequency components, \(z=\mathcal{F}(x)\), in the 2D frequency domain. The inverse, \(\mathcal{F}^{-1}(\cdot)\), restores spatial information. Given a radius threshold \(r\), low-pass filtering \(\mathrm{LPF}(z;r)\) isolates low-frequency components (LFCs), while high-pass filtering \(\mathrm{HPF}(z;r)\) captures high-frequency components (HFCs). The LFC variant \(x_{\mathrm{lfc}}\) and HFC variant \(x_{\mathrm{hfc}}\) of the original image \(x\) are reconstructed by applying \(\mathcal{F}^{-1}(\cdot)\) as follows:
\begin{equation}
\label{eq:freq}
x_{\mathrm{lfc}} = \mathcal{F}^{-1}(\mathrm{LPF}(z;r)), \quad
x_{\mathrm{hfc}} = \mathcal{F}^{-1}(\mathrm{HPF}(z;r)).   
\end{equation}
Here, \(x_{\mathrm{lfc}}\) and \(x_{\mathrm{hfc}}\) represent complementary frequency components, so that \(x \approx x_{\mathrm{lfc}} + x_{\mathrm{hfc}}\) up to numerical reconstruction error. As the radius \(r\) increases, \(x_{\mathrm{lfc}}\) contains more information, whereas \(x_{\mathrm{hfc}}\) contains less. Intuitively, \(x_{\mathrm{lfc}}\) preserves slowly varying shape and smooth structure, while \(x_{\mathrm{hfc}}\) retains sharper transitions and local details.

\ours uses this frequency interface in two places. 
During source-time training, it reconstructs source images from LFCs and uses them for QAT. 
During deployment, it applies a frequency split only to the first normalization-layer activation. Restricting frequency processing to the first normalization layer bounds runtime and memory overhead, while later normalization layers use lightweight statistic adaptation without additional frequency decomposition.
{This asymmetry — restrict to LFCs at source time, retain both bands at deployment — is the central design principle. Further analysis on distinct features of each frequency range is in Tables~\ref{tab:frequency_analysis}--\ref{tab:results_filters}.}

%%%%%%%%%%%%%%%%%%%%%%%%%%%%%%%%%%%%
%%%%%%%%%%%%%%%%%%%%%%%%%%%%%%%%%%%%
\subsection{Source-Time Low-Frequency QAT} \label{sec:method/lfc_qat}

A low-bit model has limited representational capacity compared with its full-precision counterpart. Training such a model to fit all source-domain details can allocate quantized capacity to high-frequency patterns that \textit{may not persist after deployment}. \ours therefore performs QAT on low-frequency source reconstructions.

Given labeled source data \(\mathcal{D}^{{s}}=\{(x_i^s,y_i^s)\}_{i=1}^{N}\), \ours first constructs \(x_{i,\mathrm{lfc}}^s\) using Eq.~(\ref{eq:freq}), with the training radius \(r\) fixed before deployment. It then trains the quantized CNN on \((x_{i,\mathrm{lfc}}^s,y_i^s)\) using a standard supervised QAT objective. \ours is not tied to a specific QAT algorithm; in our evaluation, it is instantiated with representative QAT methods such as LSQ~\cite{esser2020learned} and LQ~\cite{zhang2018lq}.

After training, the deployed artifact contains the fixed quantized weights, quantization parameters, affine BN parameters, and source-time BN statistics. It does not contain raw source images, labels, gradients, optimizer state, or training-time activations. Thus, LFC-QAT determines what the low-bit model learns before deployment, while preserving the deployment contract in Section~\ref{sec:deploy}.

The role of LFC-QAT is not merely to filter training images. It \textit{prepares the quantized representation} for the only adaptation mechanism available after deployment: forward-only statistic updates. Empirical evidence for this design choice, including LFC/HFC training comparisons, corruption robustness, inter-domain distance, and loss-landscape analysis, is presented later in Section~\ref{sec:analysis}.

%%%%%%%%%%%%%%%%%%%%%%%%%%%%%%%%%%%%%%%%%%%
%%%%%%%%%%%%%%%%%%%%%%%%%%%%%%%%%%%%%%%%%%%
\subsection{Deployment-Time Frequency-Aware Batch Normalization} \label{sec:method/frequency_bn}

LFC-QAT prepares the compressed model before deployment, but the target sensing environment may still differ from the source environment. \ours therefore adapts the deployed model using FABN, a forward-only update of normalization statistics from unlabeled target windows.

Let \(W_{{\tau}}=\{x_{{\tau}-k+1},\ldots,x_{{\tau}}\}\) be the current target adaptation window. During deployment, the quantized convolution and linear weights remain fixed. \ours runs \(W_{{\tau}}\) through the first quantized convolution and obtains the first BN layer's input feature \(f\). FABN applies the same frequency interface to split \(f\) into a low-frequency feature \(f_{\mathrm{lfc}}\) and a high-frequency feature \(f_{\mathrm{hfc}}\), as illustrated in Figure~\ref{fig:method_overview}. Since HFC rapidly diminishes beyond the initial layer, subsequent BN layers use lightweight statistic adaptation without additional frequency decomposition. FABN processes \(f_{\mathrm{lfc}}\) and \(f_{\mathrm{hfc}}\) differently as below.

\vspace{1ex}\noindent\textbf{Adaptation for LFCs.} 
For \(f_{\mathrm{lfc}}\), FABN initializes the BN statistics from the source domain as the running mean \(\hat{\mu}_{\mathrm{lfc},{\tau}}\) and variance \(\hat{\sigma}_{\mathrm{lfc},{\tau}}\) at test time:
\begin{equation}
\hat{\mu}_{\mathrm{lfc},0} = \hat{\mu}^{{\text{source}}}, \quad
\hat{\sigma}_{\mathrm{lfc},0} = \hat{\sigma}^{{\text{source}}},    
\end{equation}
and applies an exponential moving average (EMA) to estimate batch statistics for LFCs:
\begin{equation}
\begin{split}    
\hat{\mu}_{\mathrm{lfc},{\tau}}
=
(1-\alpha)\cdot\hat{\mu}_{\mathrm{lfc},{\tau}-1}
+
\alpha\cdot\mu_{\mathrm{lfc},{\tau}}, \\
\hat{\sigma}^{2}_{\mathrm{lfc},{\tau}}
=
(1-\alpha)\cdot\hat{\sigma}^{2}_{\mathrm{lfc},{\tau}-1}
+
\alpha\cdot\sigma^{2}_{\mathrm{lfc},{\tau}}.  
\end{split}
\end{equation}
Here, \(\mu_{\mathrm{lfc},{\tau}}\) and \(\sigma^{2}_{\mathrm{lfc},{\tau}}\) are the mean and variance of the current target-window feature \(f_{\mathrm{lfc}}\), and \(\alpha\) is the EMA momentum. This conservative update preserves source-trained low-frequency structure while allowing gradual adaptation to the target stream.

\vspace{1ex}\noindent\textbf{Adaptation for HFCs.} 
For \(f_{\mathrm{hfc}}\), FABN does not align the feature to source-domain HFC statistics. Source-time training uses low-frequency reconstructions, so source HFC statistics are not a reliable adaptation anchor. Instead, FABN estimates HFC statistics directly from the current target window:
\begin{equation}
\hat{\mu}_{\mathrm{hfc},{\tau}} = \mu_{\mathrm{hfc},{\tau}}, \quad
\hat{\sigma}^{2}_{\mathrm{hfc},{\tau}} = \sigma^{2}_{\mathrm{hfc},{\tau}}.    
\end{equation}
This lets \ours quickly capture target-specific sensing conditions such as blur, noise, texture, compression artifacts, and other local details without backpropagation or weight updates.

\vspace{1ex}\noindent\textbf{Integrating LFCs and HFCs.} 
After computing the individual statistics for \(f_{\mathrm{lfc}}\) and \(f_{\mathrm{hfc}}\), FABN combines them to produce the final BN parameters \(\hat{\mu}_{{\tau}}\) and \(\hat{\sigma}_{{\tau}}\):
\begin{equation}  \label{eq:fabn_combine}
\hat{\mu}_{{\tau}} = \hat{\mu}_{\mathrm{lfc},{\tau}} + \hat{\mu}_{\mathrm{hfc},{\tau}}, \quad
\hat{\sigma}^{2}_{{\tau}}
=
\hat{\sigma}^{2}_{\mathrm{lfc},{\tau}}
+
\hat{\sigma}^{2}_{\mathrm{hfc},{\tau}}.
\end{equation}
Eq.~(\ref{eq:fabn_combine}) uses a covariance-free approximation for efficiency. This is appropriate for \ours's deployment setting because the split is applied only at the first BN input using complementary frequency masks, and the goal is lightweight forward-only statistic adaptation rather than exact reconstruction of full activation covariance. %We evaluate the effect of this approximation in Section~\ref{sec:analysis}.

The first BN layer then normalizes \(f\) using \(\hat{\mu}_{{\tau}}\) and \(\hat{\sigma}^{2}_{{\tau}}\), while keeping the affine BN parameters fixed. \ours does not update convolution weights, linear weights, BN affine parameters, or quantization parameters during deployment. For later BN layers, \ours avoids further frequency decomposition and uses lightweight statistic adaptation~\cite{nado2020evaluating,schneider2020robustness}, keeping the runtime overhead bounded.

%%%%%%%%%%%%%%%%%%%%%%%%%%%%
%%%%%%%%%%%%%%%%%%%%%%%%%%%%
\subsection{Deployment State and Complexity}
\label{subsec:deployment_state}

\ours adds one frequency split at the first normalization-layer input. If the first activation has shape \(C_1 \times H_1 \times W_1\), where $C$, $H$, $W$ denote the number of channels, height and width, an FFT-based implementation costs \(O(C_1H_1W_1\log(H_1W_1))\) for this layer only. Later layers perform only lightweight statistic updates without frequency decomposition. Section~\ref{sec:implementation} describes the deployed artifact and mutable runtime state.

\section{Implementation}
\label{sec:implementation}

This section describes how \ours is instantiated under the adaptive quantized inference contract in Section~\ref{sec:deploy}. We focus on the deployed artifact, quantized runtime, adaptation configuration, and measurement protocol; detailed hyperparameters are in  Appendix~\ref{app:reproducibility}.

%%%%%%%%%%%%%%%%%%%%%%%%%%%%%%%%%%%%%%%%
%%%%%%%%%%%%%%%%%%%%%%%%%%%%%%%%%%%%%%%%
\subsection{Deployed Runtime and Mutable State}
\label{subsec:deployed_runtime}

The deployed \ours artifact contains fixed quantized convolution and linear weights, quantization parameters, affine BN parameters, source-time BN statistics, and FABN statistic buffers. It does not contain source images, target labels, gradients, optimizer state, training activations, or mutable full-precision weights. During deployment, the only mutable state is the inference state \(s_{{\tau}}\), consisting of normalization statistics and bounded statistic buffers. FABN updates this state using forward passes over the unlabeled target window \(W_{{\tau}}\) without backpropagation, optimizer update, or weight update.

For hardware deployment, we use PyTorch \texttt{torch.ao.quantization} with the QNNPACK backend. Convolution and linear layers are wrapped with \texttt{QuantStub}/\texttt{DeQuantStub}, prepared, and converted using \texttt{torch.ao.quantization}. {The weight and activation ranges including learned step sizes are fixed at training time --- no post-training calibration pass--- and are exported directly to integer by prepare/convert workflow.} BN layers are {not fused to CNNs and} excluded from quantization, because \ours requires explicit mutable BN/FABN statistics during test-time adaptation. Thus, convolution and linear layers execute through quantized kernels, while BN/FABN remains an explicit inference module whose statistics can be updated online. 

Our edge-device measurements report quantized execution for the compute-heavy convolution and linear layers. The 2-bit and 4-bit experiments evaluate low-bit robustness and model-size behavior; native hardware acceleration at these bit-widths depends on backend support and is not claimed. % unless explicitly stated.

\vspace{-1ex}

%%%%%%%%%%%%%%%%%%%%%%%%%%%%%%%%%%%%
%%%%%%%%%%%%%%%%%%%%%%%%%%%%%%%%%%%%
\subsection{Training and Adaptation Configuration}
\label{subsec:training_adaptation_config}

We use pre-activation ResNet models for the main evaluation: ResNet26 for CIFAR10 and ResNet18/50 for ImageNet{, adapted on CIFAR10-C (severity 5) and ImageNet-C (severity 3), respectively}. All models are trained from scratch. CIFAR10 experiments use three random seeds, while ImageNet experiments use one seed because of computational cost. CIFAR10 training is performed on a single NVIDIA GeForce RTX 3090 GPU, and ImageNet training is performed on four NVIDIA A100 GPUs.

\ours is compatible with standard QAT methods. We instantiate it with LSQ and LQ, following their official implementations and reported hyperparameters. Unless otherwise specified, LSQ and LQ use 2-bit quantization for both weights and activations in the main low-bit experiments. For model-size and accuracy--latency comparisons, we also evaluate 4-bit and 8-bit LSQ models on ResNet18 and ResNet50. Full-precision models use the same architecture and data pipeline without quantization. For QAT baselines, FFC-trained models use original full-frequency source images, while \ours uses low-frequency source reconstructions for LFC-QAT.

For LFC-QAT, we use radius \(r=8\) for CIFAR10 and \(r=56\) for ImageNet. {Both radii correspond to one-quarter of the input side length and are validated by the frequency-radius analysis in Section~\ref{sec:analysis}, which evaluates the trade-off between clean accuracy and corruption robustness}{: larger radii improve clean accuracy, whereas corruption robustness peaks at a smaller radius (Tables~\ref{tab:frequency_analysis},~\ref{tab:frequency_analysis_c}). In particular, r=28 slightly outperforms r=56 on ImageNet-C, while r=56 yields higher clean accuracy. We retain r=56 as a consistent resolution-scaled default rather than selecting corruption-accuracy optimum for each dataset.} At deployment time, FABN operates on the current unlabeled target window \(W_{{\tau}}\). The main robustness evaluation uses target window size \(k=64\) and 4 to observe \ours's behavior on both a large and small window size, respectively, and the more detailed batch/window-size sensitivity study uses \(k \in \{1,2,4,8,16,32,64\}\). The EMA momentum for LFC statistics is \(\alpha=0.1\). {Further implementation details are in Appendix~\ref{app:reproducibility}}.

FABN is applied only at the first BN layer by default; later BN layers use lightweight statistic adaptation without frequency decomposition. {Specifically, FABN splits LFCs and HFCs using the same radius as source-time LFC QAT.} We denote by CoDA$^\ast$ the variant that applies FABN to all BN layers and use it only for ablation and overhead analysis. For episodic adaptation, BN/FABN statistics are reset at the beginning of each target domain. For continual adaptation, statistics are carried forward across the target stream unless otherwise specified. No target labels or source samples are used in either setting.

\begin{figure}[t]
    % \centering

    \includegraphics[height=0.35\linewidth]{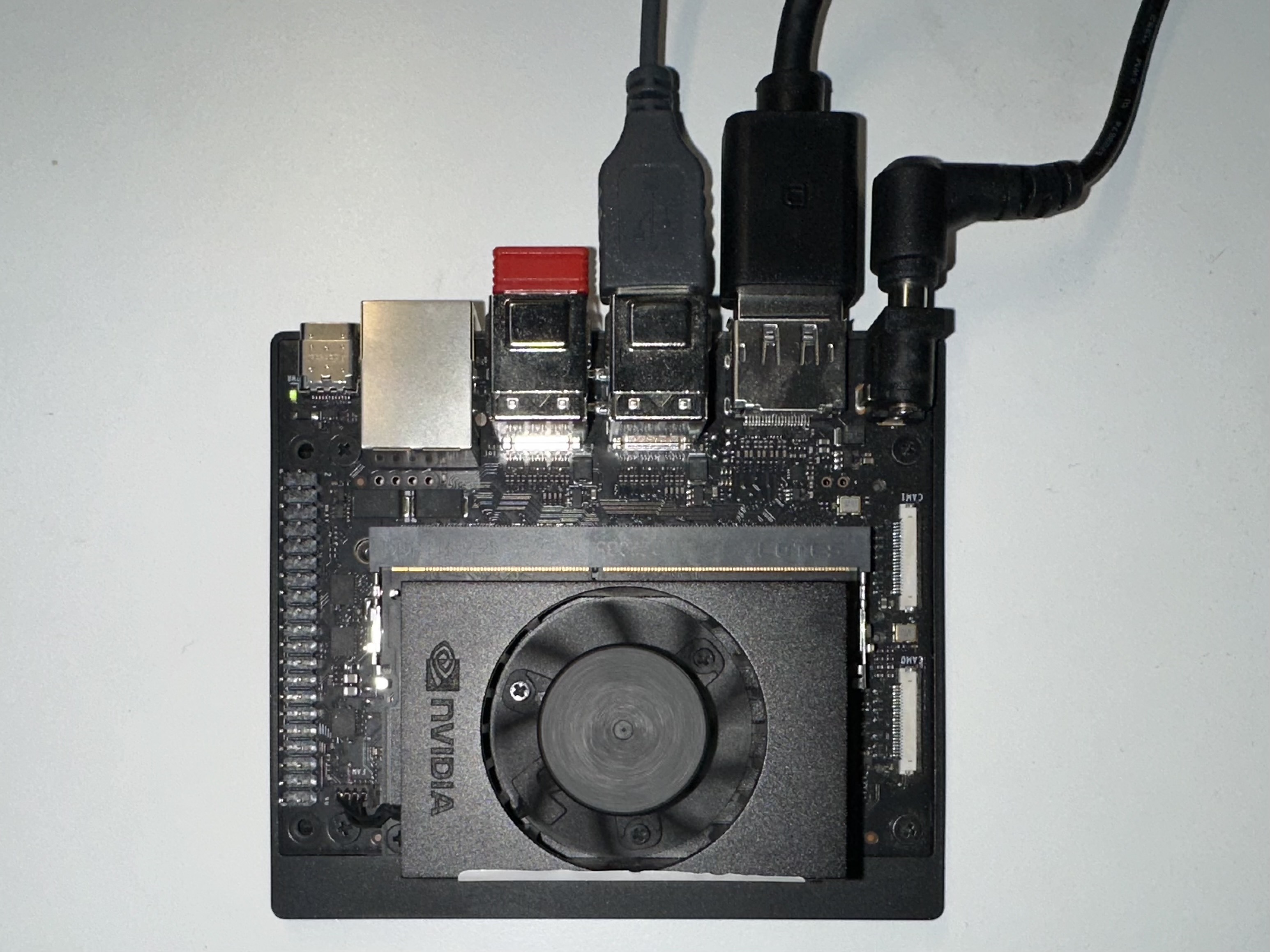}
    % \hfill 
    \includegraphics[height=0.35\linewidth]{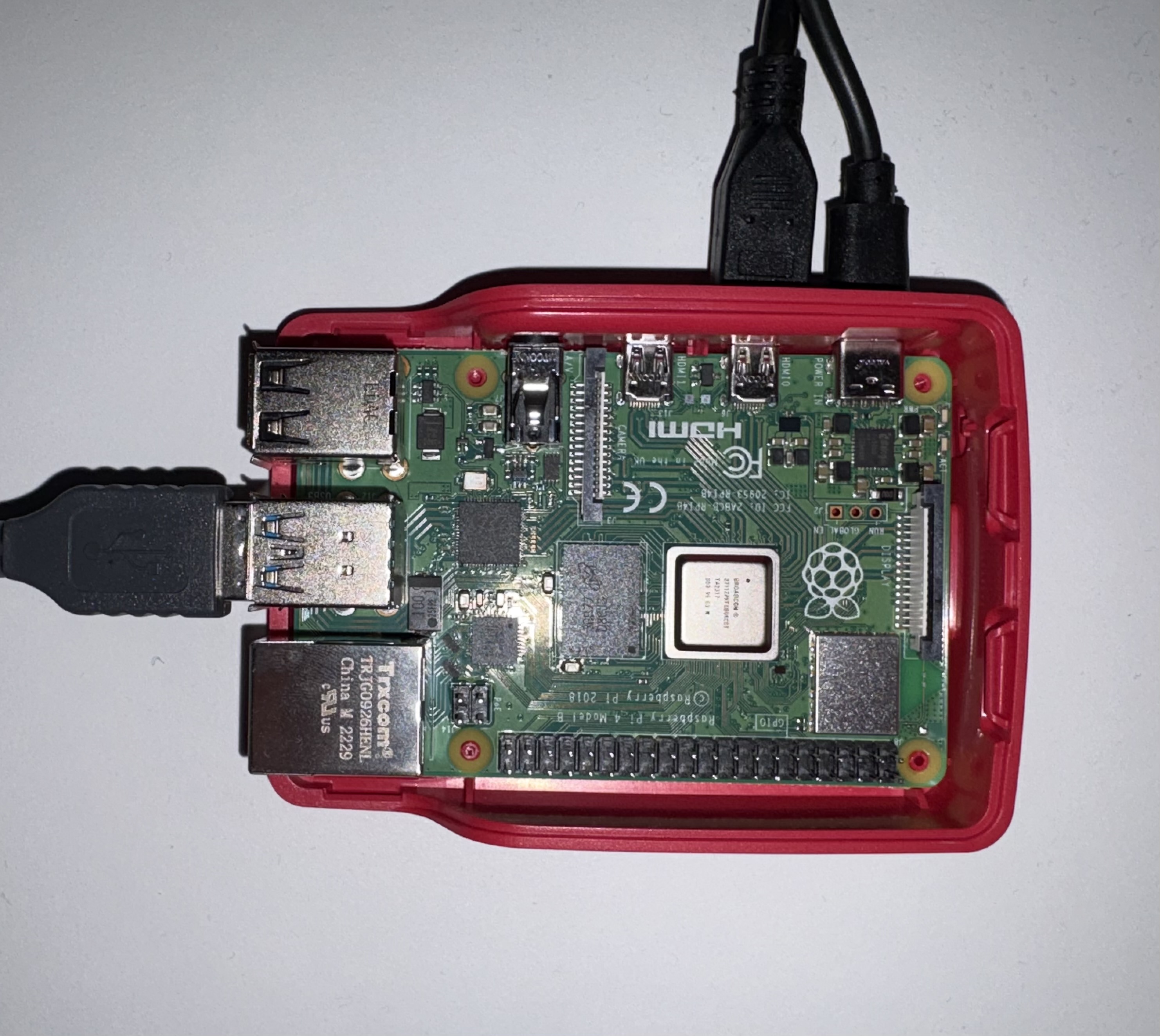}

    \vspace{-2ex}
    \captionof{figure}{
    Edge platforms: Jetson Orin Nano (left) and Raspberry Pi 4B (right).
    }
    \Description{Two photographs of the edge devices used in the experiments: an NVIDIA Jetson Orin Nano developer board on the left and a Raspberry Pi 4 Model B board on the right.}
    \vspace{-4ex}
    \label{fig:jetson_photo}
\end{figure}

\begin{table*}[t]
    \centering
    \caption{Main results of \ours (\%) compared to the forward-only baseline TTA (NORM, TEMA) on ImageNet-C with $k=64$ under various settings, including two QATs (LSQ and LQ) with three bit-widths (2, 4, and 8 bits) and DeepAug (marked as ``+A'').
    }
    \vspace{-2ex}
    \resizebox{0.85\linewidth}{!}{
    {\fontsize{9pt}{10pt}\selectfont
    \setlength{\tabcolsep}{1.7mm}   % 기본값은 6pt 정도
    
    \begin{tabular}{ll|cc|ccc|ccc|ccc}   
        \toprule
        \multirow{2}{*}{Model} & \multirow{2}{*}{Method} & \multicolumn{2}{c|}{32-bit} & \multicolumn{3}{c|}{2-bit} & \multicolumn{3}{c|}{4-bit} & \multicolumn{3}{c}{8-bit} \\
        \cmidrule{3-13}
        & & FP & FP+A & LSQ & LSQ+A & LQ & LSQ & LSQ+A & LQ& LSQ & LSQ+A & LQ \\
        \midrule
        \multirow{5}{*}{ResNet18} & w/o TTA & 
            22.36 & 41.29 & 16.25 & 27.74 & 18.66 & 19.90 & 32.46 & 20.89 & 20.15 & 32.59 & 21.11\\
        & NORM &
            29.13 & 49.82 & 24.44 & 36.72 & 28.35 & 28.50 & 40.41 & 28.68 & 29.23 & 42.10 & 30.42 \\
        & TEMA &
            29.33 & 48.33 & 21.78 & 32.61 & 22.91 & 27.04 & 39.31 & 28.41 & 25.36 & 36.72 & 29.07 \\
        % \midrule
        \cmidrule{2-13}
        & \ours & 
            \textbf{33.80} & \textbf{54.62} & \textbf{31.43} & \textbf{42.13} & \textbf{34.27}	& \textbf{33.50} & \textbf{45.92} & \textbf{34.46}	&  \textbf{34.17}	& \textbf{46.75} &\textbf{34.48} \\

        & (ours)  & 
            (+4.47) & (+4.80) & (+6.99) & (+5.41) & (+5.92)	& (+5.00)	& (+5.51) & (+5.78)	& (+4.94) & (+4.65) &(+4.06)  \\
        \midrule
        \multirow{5}{*}{ResNet50} & w/o TTA & 
            27.74 & 51.91 & 19.76 & 32.97 & 22.23 & 21.57 & 39.36 & 22.96 &
            22.58 & 40.77 & 15.70 \\

        & NORM &
            36.20 & 58.59 & 30.44 & 42.53 & 32.65 & 31.30 & 47.78 & 31.93 & 31.87 & 48.39 & 26.52 \\
        & TEMA &
            34.96 & 51.03 & 28.20 & 39.36 & 33.00 & 29.20 & 44.12 & 31.04 & 26.83 & 41.89 & 29.42 \\
        \cmidrule{2-13}
        & \ours& 
            \textbf{40.73} & \textbf{61.38}	& \textbf{35.22}	& \textbf{46.74} &  \textbf{38.01}	& \textbf{36.25}	& 
            \textbf{52.86} & 
            \textbf{36.49}	& \textbf{36.68}	& 
            \textbf{54.19} & 
            \textbf{35.02} \\

        & (ours)  & 
            (+4.53) & (+2.79)	& (+4.78)	& (+4.21) &  (+5.01)	& (+4.95)	& (+5.08) & (+4.56)	& (+4.81)	& (+5.80) & (+5.60) \\
        \bottomrule
    \end{tabular}
    }
    }
    \vspace{-1ex}
    \label{tab:results_imagenet}
\end{table*}

%%%%%%%%%%%%%%%%%%%%%%%%%%%%%%%%%%%%
%%%%%%%%%%%%%%%%%%%%%%%%%%%%%%%%%%%%
\vspace{-1ex}

\subsection{Edge Platforms and Measurement Protocol}
\label{subsec:edge_measurement_protocol}

We evaluate systems performance on Raspberry Pi 4B and NVIDIA Jetson Orin Nano CPU (Figure~\ref{fig:jetson_photo}). We conduct all edge deployment experiments under PyTorch 2.2.0. Our primary platform is an NVIDIA Jetson Orin Nano, where we run CPU inference using 4 out of 6 CPU cores. The device runs JetPack 6.0 (L4T 36.3.0). Additionally, we evaluate CPU inference on Raspberry Pi utilizing 2 out of 4 CPU cores running Raspberry Pi OS (64-bit). Both platforms utilize QNNPACK backend for quantized execution. 
Unless otherwise specified, GPU acceleration is disabled for both measurements so that the reported results reflect CPU-side quantized inference.

We additionally evaluate on a Samsung laptop CPU reference platform, equipped with Intel i5-1035G4 CPU, Windows 11, PyTorch 2.2.0, and FBGEMM backend for quantized execution. We used 4 out of 8 CPU cores. This platform tests whether \ours's system behavior generalizes beyond edge platforms.

We measure latency, peak memory and CPU utilization. Unless otherwise specified, the main edge-device measurements use window size 16. For latency, we perform 100 warm-up runs and report the average wall-clock time over 100 inference/adaptation runs. For peak memory, we perform 100 warm-up runs and record peak memory over 100 runs at 0.1-second intervals utilizing \texttt{psutil} package. For CPU utilization, we monitor processor utilization using \texttt{psutil} package across both platforms.

All timing measurements include both the forward pass and the TTA step. For forward-only baselines such as NORM and TEMA, measurements include the required forward passes and statistic updates. For \ours, measurements include the first-layer frequency split, FABN statistic update, and remaining quantized forward pass. For backpropagation-based baselines such as TENT and SAR, measurements include the backward pass and optimizer update.

%%%%%%%%%%%%%%%%%%%%%%%%%%%%%%%%%%%%
%%%%%%%%%%%%%%%%%%%%%%%%%%%%%%%%%%%%
\vspace{-1ex}
\subsection{Baselines and Artifacts}
\label{subsec:baselines_reproducibility}

We separate deployment-compatible baselines from accuracy-oriented upper bounds. NORM and TEMA are forward-only baselines and can be evaluated under quantized inference because they do not require weight updates. TENT, SAR, and CoTTA require backpropagation or parameter updates. Since backpropagation through low-bit integer kernels is not supported in our inference-only deployment path, we evaluate these methods by reusing the corresponding QAT-trained weights and dequantizing them to full precision for adaptation. These methods therefore serve as contract-violating upper bounds rather than deployment-compatible competitors.

All baseline models are trained on full-frequency source images unless otherwise specified. \ours uses LFC-QAT during source-time training and FABN during deployment-time adaptation. For augmentation compatibility experiments such as DeepAug, we apply the augmentation during source-time training and use the same deployment protocol for adaptation.
%
% We will release training scripts for FP, LSQ, LQ, and \ours, frequency reconstruction code and FABN implementation.
Full training details for reproducability are listed in Appendix~\ref{app:reproducibility}.

%%%%%%%%%% Evaluation
\section{Experiments: Accuracy under Domain Shift}\label{sec:evalutation}

This section evaluates whether \ours improves target-domain accuracy under the deployment contract in Section~\ref{sec:deploy}. Section~\ref{sec:system} evaluates the corresponding system cost, including latency, memory, and CPU utilization, and Section~\ref{sec:analysis} analyzes individual design choices.

\vspace{-1ex}

%%%%%%%%%%%%%%%%%%%%%%%%%%%%%
%%%%%%%%%%%%%%%%%%%%%%%%%%%%%
\subsection{Evaluation Overview}
\label{subsec:accuracy_overview}

We evaluate \ours on four domain-shift benchmarks {with an additional real-world environmental shift benchmark}. 
CIFAR10-C~\cite{hendrycksbenchmarking} is used as target data for models trained on CIFAR10~\cite{krizhevsky2009learning}, while ImageNet-C~\cite{hendrycksbenchmarking}, ImageNet-R~\cite{hendrycks2021many}, and ImageNet-Sketch~\cite{wang2019learning} are used as target data for models trained on ImageNet. 
CIFAR10-C and ImageNet-C, {which are the standard suite of the TTA literature~\cite{nado2020evaluating, su2024unraveling, wang2020tent, niu2023towards},} evaluate common corruptions; ImageNet-R/Sketch evaluate rendition and sketch-style shifts. 
{ImageNet-ES-Diverse~\cite{baek2025adaptive} further evaluates real capture-condition (lighting) shifts using physically captured images under Auto-Exposure and varying lighting.}
All target samples are unlabeled and no source samples are used during adaptation.

We evaluate ResNet backbones~\cite{he2016identity} for the main results and MobileNetV3/EfficientNet~\cite{howard2019mobilenetv3,tan2019efficientnet} for lightweight-model settings. 
\ours is instantiated with representative QAT methods, LSQ~\cite{esser2020learned} and LQ~\cite{zhang2018lq}, over multiple bit-widths. We separate baselines into two groups. NORM~\cite{nado2020evaluating,schneider2020robustness} and TEMA~\cite{su2024unraveling} are deployment-compatible forward-only baselines because they do not require weight updates. 
TENT~\cite{wang2020tent}, SAR~\cite{niu2023towards}, and CoTTA~\cite{wang2022continual} require backpropagation or parameter updates with full-precision (FP) models; we report them as accuracy-oriented upper bounds rather than deployment-compatible competitors. All baseline models are trained on full-frequency source images unless otherwise specified, while \ours uses LFC-QAT during source-time training and FABN during deployment-time adaptation.
Lastly, we utilize DeepAug~\cite{hendrycks2021many} to evaluate compatibility between LFC-selectivity and existing augmentations.

%%%%%%%%%%%%%%%%%%%%%%%%%%%%%%%%
%%%%%%%%%%%%%%%%%%%%%%%%%%%%%%%%
\vspace{-1ex}

\subsection{Main Corruption and Domain-Shift Results}

% %ImageNet-R/-Sketch
% Additional evaluations on CIFAR10-C (Table~\ref{tab:results_cifar}), ImageNet-R, and ImageNet-Sketch (Table~\ref{tab:results_imagenetrsketch}) confirm that \ours consistently maintains strong performance across diverse datasets including other types of domain shifts. 
% %
% For comparison, we also include results of TENT and SAR applied to quantized models, though their weights are fully de-quantized at inference (no compression impact). Under identical weight conditions, \ours's frequency-asymmetric, forward-only adaptation consistently proves more effective than parameter updates based on back-propagation, reinforcing its superiority in resource-constrained scenarios. 

Tables~\ref{tab:results_imagenet}, \ref{tab:results_cifar}, and \ref{tab:results_imagenetrsketch} show TTA performance on ImageNet-C, CIFAR10-C, and ImageNet-R/-Sketch datasets, respectively. While ImageNet-C and CIFAR10-C are standard TTA benchmarks, ImageNet-R/-Sketch are additionally utilized for evaluating generalizability of our approach. 

\vspace{1ex}\noindent\textbf{ImageNet-C.} 
Table~\ref{tab:results_imagenet} reports ImageNet-C accuracy across ResNet18/50, QAT methods, bit-widths, and DeepAug settings. \ours improves over the forward-only baselines across the main quantized settings. Notably, on ResNet-18, \ours even outperforms full-precision models with TTA by \textbf{up to 5.35 percentage points (pp)}, while \textbf{reducing model size by 4 to 16 times}.   
This result shows that post-hoc BN-statistic adaptation alone is not sufficient for low-bit deployment: the quantized representation must also be prepared before deployment. \ours achieves this by pairing LFC-QAT with FABN, improving robustness while keeping the deployed convolution and linear layers fixed and quantized.

\begin{table}[t]
    \centering
    \caption{Compatibility between \ours and back-propagation-based TTAs on ImageNet-C under FP models ($k=64$).
    }
    \vspace{-2ex}
    \resizebox{\linewidth}{!}{
    \begin{tabular}{l|l|cc|cc}
        \toprule
        Model & Aug. & TENT & TENT+\ours & SAR & SAR+\ours \\
        \midrule
        \multirow{2}{*}{ResNet18} & None      & 30.6 & \textbf{34.0} (+3.4) & 33.9 & \textbf{34.8} (+0.9)\\
                 & DeepAug & 50.7 & \textbf{54.3} (+3.6) & \textbf{54.2} & 53.8 (-0.4) \\
        \midrule
        ResNet50 & DeepAug & 59.2 & \textbf{61.7} (+2.5) & 57.1 & \textbf{61.9} (+4.8) \\
        \bottomrule
    \end{tabular}
    }
    \vspace{-2ex}
    \label{tab:results_imagenet_TENTSAR}
\end{table}

\begin{table}[t]
    \centering
    % \centering    
    \caption{Classification acc. (\%) of \ours compared to the baseline TTAs and QATs on ImageNet-C with  ResNet50 and $k=4$.}
    % \vspace{-3ex}
    \vspace{-2ex}
    \resizebox{\linewidth}{!}{
    \begin{tabular}{l|c|cc|cc|cc}
    \toprule
    \multirow{2}{*}{Method} & 32-bit& \multicolumn{2}{c|}{2-bit} & \multicolumn{2}{c|}{4-bit} & \multicolumn{2}{c}{8-bit} \\
    \cmidrule{2-8}
     & FP & LSQ & LQ & LSQ & LQ & LSQ & LQ \\
    \midrule
    w/o TTA & 27.74 & 19.76 & 22.23 & 21.57 & 22.96 & 22.58 & 15.70 \\
    NORM & 26.36 & 20.78 & 22.79 & 22.31 & 23.48 & 23.14 & 18.03 \\
    TEMA & 30.12 & 23.86 & 27.51 & 25.60 & 27.06 & 24.38 & 27.19 \\
    TENT & 15.14 & - & - & - & - & - & - \\
    SAR & 14.53 & - & - & - & - & - & - \\
    \midrule
    \ours & \textbf{36.01} & \textbf{30.03} & \textbf{32.87} & \textbf{31.48} & \textbf{32.19}& \textbf{31.63} & \textbf{31.72} \\
    (ours) & (+5.89) & (+6.17) & (+5.36) & (+5.88) & (+5.13) &  (+7.25) & (+4.53) \\
    \bottomrule
    \end{tabular}
    }
    \vspace{-2ex}
    \label{tab:results_image4}
\end{table}

\ours also remains compatible with DeepAug when the augmentation is applied during source-time training. This suggests that LFC-QAT does not replace existing robustness-oriented training augmentations; it can be combined with them. 
Table~\ref{tab:results_imagenet_TENTSAR} reports an additional full-precision (FP) compatibility study with TENT and SAR. The result shows that \ours is generally compatible with backpropagation-based TTA upper bounds, improving most tested settings. Because these methods require backpropagation or parameter updates, we interpret this result only as a compatibility study, not as the main deployment comparison.

Table~\ref{tab:results_image4} reports \ours's robustness on a small window size ($k=4$), regarding a practical on-device TTA concern: edge deployments may not have access to large, balanced target batches~\cite{danilowski2026botta}. 
%
%To manage computational overhead of small batch sizes, we evaluate on a randomly selected 50\% subset (5,000 samples) per corruption under a single light setting.
Both TENT and SAR experience significant performance drops even with FP model and backpropagation, which supports the concern that backpropagation-based TTA can be unstable under small target windows.
In contrast, \ours is more stable than the compared TTA baselines under small windows (outperforming with large margins) because its LFC statistics are updated conservatively from a source-initialized state, while HFC statistics are estimated directly from the current target window. This dual-track update allows \ours to adapt with limited target samples without using labels, source data, gradients, optimizer state, or weight updates.

\begin{table}[t]
    \centering
    % \centering    
    \caption{Classification acc. (\%) of \ours compared to the baseline TTAs and QATs on CIFAR10-C using ResNet26 ($k=64$).}
    \vspace{-2ex}
    \resizebox{\linewidth}{!}{
    \begin{tabular}{l|c|cc|cc|cc}
    \toprule
    \multirow{2}{*}{Method} & 32-bit& \multicolumn{2}{c|}{2-bit} & \multicolumn{2}{c|}{4-bit} & \multicolumn{2}{c}{8-bit} \\
    \cmidrule{2-8}
     & FP & LSQ & LQ & LSQ & LQ & LSQ & LQ \\
    \midrule
    w/o TTA & 53.86 & 50.15 & 54.35 & 52.65 & 55.40 & 51.60 & 53.16 \\
    NORM & 68.42 & 65.52 & 69.29 & 64.23 & 68.05 & 64.23 & 67.47 \\
    TEMA & 53.97 & 39.62 & 50.11 & 40.50 & 52.35 & 42.30 & 52.71 \\
    TENT & 72.64 & - & - & - & - & - & - \\
    SAR & 71.66 & - & - & - & - & - & - \\
    %TENT & 72.64 & 70.18 & 73.72 & 70.43 & 73.15 & 71.09 & 72.90 \\
    %SAR & 71.66 & 68.99 & 73.60 & 69.46 & 73.16 & 69.52 & 73.06 \\
    \midrule
    \ours & \textbf{73.05} & \textbf{71.22} & \textbf{76.15} & \textbf{72.50} & \textbf{76.38 }& \textbf{72.86} & \textbf{75.52} \\
    (ours) & (+0.41) & (+5.70) & (+6.86) & (+8.27) & (+8.33) & (+8.63) & (+8.05) \\  
    \bottomrule
    \end{tabular}
    }
    \vspace{-2ex}
    \label{tab:results_cifar}
\end{table}

\begin{figure}[t]
  \centering 
    \includegraphics[width=.8\linewidth]{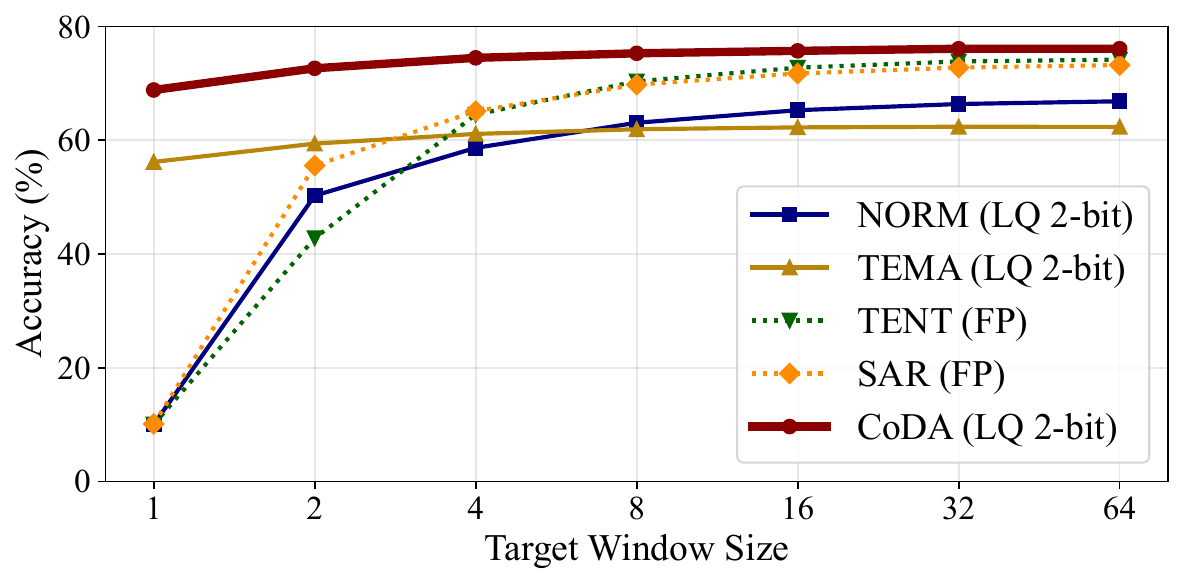} 
    \vspace{-2.5ex}
    \caption{Classification accuracy (\%) comparison under different window sizes with ResNet26 on CIFAR10-C.}
    \Description{Line plot of classification accuracy versus target window size (1, 2, 4, 8, 16, 32, 64) on CIFAR10-C with ResNet26. CoDA with 2-bit LQ is the highest curve at every window size, starting near 69 percent at size 1 and rising to about 77 percent. NORM, TENT, and SAR start near 10 percent at size 1 and climb toward CoDA as the window grows, while TEMA stays flat around 56 to 62 percent.}
    \label{fig:batch_size}
    \vspace{-2ex}
\end{figure}

\vspace{1ex}\noindent\textbf{CIFAR10-C.} 
Table~\ref{tab:results_cifar} reports CIFAR10-C results on ResNet26. \ours significantly improves over NORM and TEMA across LSQ, LQ and FP settings and remains competitive with or better than backpropagation-based FP upper bounds while using only forward-pass statistic adaptation. These results are consistent with ImageNet-C: \ours is most beneficial when the model is compressed and must adapt without updating weights.
Figure~\ref{fig:batch_size} evaluates the effect of target adaptation window size on CIFAR10-C, verifying its robustness under smaller window sizes. In contrast, NORM, TENT, and SAR experience rapid performance degradation as the window size falls below 16.

\begin{table}[t]
    \centering
    \caption{Classification acc. (\%) of ResNet18/50 on ImageNet-R/-Sketch quantized in 2-bit with and w/o TTA.
    }
    \vspace{-2ex}
    \resizebox{\linewidth}{!}{
    \begin{tabular}{l|l|cc|cc|cc|cc}
        \toprule
        \multirow{3}{*}{Dataset} &\multirow{3}{*}{Method} & \multicolumn{4}{c|}{$k=4$} & \multicolumn{4}{c}{$k=64$} \\
        \cmidrule{3-10}
        & & \multicolumn{2}{c|}{ResNet18} & \multicolumn{2}{c|}{ResNet50} & \multicolumn{2}{c|}{ResNet18} & \multicolumn{2}{c}{ResNet50} \\
        \cmidrule{3-10}
        & & LSQ & LQ & LSQ & LQ & LSQ & LQ & LSQ & LQ \\
        \midrule
        \multirow{4}{*}{\makecell[c]{ImageNet-\\R}} & 
        w/o TTA 
        & 19.6 & 22.0 & 21.8 & 24.9 & 19.6 & 22.0 & 21.8 & 24.9 \\
        & NORM 
        & 15.6 & 17.6 & 18.3 & 21.3 & 21.7 & 24.2 & 23.7 & 27.2 \\
        & TEMA 
        & 17.6 & 20.3 & 21.1 & 24.8 & 18.3 & 19.4 & 22.0 & 25.3 \\
        % TENT &
        % 20.5 & 23.6 & 23.3 & 26.8 & 10.2         & 11.7 & 12.7         & 15.3 \\
        % SAR & 
        % 20.9 & 23.6 & 23.3 & 26.7 & 10.8         & 11.7 & 13.0         & 15.3 \\
        \cmidrule{2-10}
        & \ours 
        & \textbf{19.9} & \textbf{22.8} & \textbf{21.9} & \textbf{25.5} & \textbf{21.9} & \textbf{24.6} & \textbf{24.0} & \textbf{27.8} \\
        \midrule
        \multirow{4}{*}{\makecell[c]{ImageNet-\\Sketch}} & 
        w/o TTA 
        & \textbf{9.6}  & 11.5 & 10.6 & 13.6 & 9.6  & 11.5 & 10.6 & 13.6 \\
        & NORM 
        & 7.5  & 8.5 & 8.9  & 11.3 & \textbf{11.4} & 12.9 & 12.6 & 16.0 \\
        & TEMA 
        & 9.1 & 11.3 & \textbf{11.8} & \textbf{15.1} & 10.0 & 11.9 & 12.7 & \textbf{16.2} \\
        % TENT &
        % 20.5 & 23.6 & 23.3 & 26.8 & 10.2         & 11.7 & 12.7         & 15.3 \\
        % SAR & 
        % 20.9 & 23.6 & 23.3 & 26.7 & 10.8         & 11.7 & 13.0         & 15.3 \\
        \cmidrule{2-10}
        & \ours 
        & 8.8 & \textbf{11.6} & 11.5 & 14.7 & 9.9 & \textbf{13.0} & \textbf{13.3} & \textbf{16.2} \\
        \bottomrule
    \end{tabular}
    }
    \vspace{-2ex}
    \label{tab:results_imagenetrsketch}
\end{table}

\vspace{1ex}\noindent\textbf{ImageNet-R and ImageNet-Sketch.} 
Table~\ref{tab:results_imagenetrsketch} evaluates shifts beyond standard synthetic corruptions on ImageNet-R and ImageNet-Sketch. %ImageNet-ES Diverse is particularly challenging because real-world environment and sensor shifts can differ from the commonly used synthetic corruptions. %In this setting, existing TTA methods sometimes provide limited gains, and in several cases no adaptation is competitive.
%\ours provides the strongest performance among the tested methods, indicating that frequency-asymmetric adaptation is not limited to standard corruption benchmarks.
%
\ours remains effective on most settings and robust to the small window size ($k=4$), although the gains are smaller than on corruption benchmarks {as they are for the forward-only baselines, with \ours on par with or ahead of them on average}. We therefore treat these results as evidence of generality rather than as the primary deployment claim. The main result is that \ours improves robustness in the settings most aligned with adaptive quantized edge inference: compressed CNNs under unlabeled corruption and sensing shifts.

\subsection{{Evaluation on Environmental-shifts for Edge-sensing}} 
\label{app:env_shift}
{As synthetic corruptions may favor a frequency-based design, to further evaluate \ours's robustness on more diverse and realistic edge sensing shifts, we utilize ImageNet-ES-Diverse~\cite{baek2025adaptive}, which captures real photographs through a physical camera under various sensor parameters and lighting conditions. Since ~\ours adapts the model rather than the camera, we use the Auto-Exposure captures rather than the manual sensor-parameter sweep, so lighting condition is the only shift variable. This isolates lighting-induced capture shifts, excluding temporal dynamics such as streaming or robotic deployment. Each of the six lighting conditions (L1--L4, L6, L7, following the dataset) is treated as a target domain, comparing no adaptation, NORM, TEMA, and CoDA quantized to 2/4/8-bits using LSQ.
Each light setting is defined by combination of magnitude and direction from the dual (left/right) lights. As Table~\ref{tab:imagenet-diverse} shows, 
while TEMA outperforms its full-precision baseline at lower bit-widths, \ours achieves the best accuracy at full-precision, 8-, and 4-bit, and competitive second-best at 2-bit, trailing TEMA by a small margin (15.42 vs. 15.83). This suggests that \ours's robustness gains extend from synthetic corruption benchmarks to real capture-condition shifts of a physical camera.}

\begin{table}[t]
\centering
\caption{{Classification acc. (\%) of \ours compared to the baseline TTAs on ImageNet-ES-Diverse using ResNet50 (k=64).}}
\vspace{-2ex}
\label{tab:imagenet-diverse}
\setlength{\tabcolsep}{2pt} 
    \begin{tabular}{clccccccc}
    \toprule
    Bits. & Method & L1 & L2 & L3 & L4 & L6 & L7 & Avg. \\
    \midrule
    \multirow{4}{*}{32} & w/o TTA & \underline{14.68} & \underline{13.70} & \underline{14.04} & \underline{14.28} & \underline{12.46} & 11.60 & \underline{13.46} \\
     & NORM    & 13.12 & 13.32 & 13.50 & 13.26 & 12.30 & \underline{11.76} & 12.88 \\
     & TEMA    & 12.29 & 12.45 & 12.45 & 12.11 & 10.71 & 10.00 & 11.67 \\
     \cmidrule{2-9}
     & CoDA    & \textbf{15.40} & \textbf{14.22} & \textbf{15.64} & \textbf{14.98} & \textbf{13.54} & \textbf{12.50} & \textbf{14.38} \\
    \midrule
    \multirow{4}{*}{8} & w/o TTA & 11.12 & 10.66 & 11.16 & 11.04 & 9.46  & 9.04  & 10.41 \\
     & NORM    & 12.44 & 11.86 & 12.74 & 12.08 & 11.12 & 10.64 & 11.81 \\
     & TEMA    & \underline{13.44} & \underline{14.04} & \underline{14.19} & \underline{14.41} & \underline{12.16} & \underline{11.69} & \underline{13.32} \\
     \cmidrule{2-9}
     & CoDA    & \textbf{15.82} & \textbf{15.34} & \textbf{16.88} & \textbf{16.18} & \textbf{13.90} & \textbf{13.60} & \textbf{15.29} \\
    \midrule
    \multirow{4}{*}{4} & w/o TTA & 11.80 & 11.68 & 12.24 & 11.98 & 10.48 & 9.64  & 11.30 \\
     & NORM    & 12.12 & 11.88 & 12.72 & 12.48 & 10.80 & 10.40 & 11.73 \\
     & TEMA    & \underline{15.60} & \underline{15.28} & \underline{14.88} & \underline{15.34} & \underline{13.58} & \underline{12.92} & \underline{14.60} \\
     \cmidrule{2-9}
     & CoDA    & \textbf{16.30} & \textbf{16.12} & \textbf{16.92} & \textbf{16.62} & \textbf{14.74} & \textbf{13.72} & \textbf{15.74} \\
    \midrule
    \multirow{4}{*}{2} & w/o TTA & 11.40 & 11.50 & 11.74 & 11.40 & 10.00 & 9.76  & 10.97 \\
     & NORM    & 10.14 & 9.82  & 9.84  & 10.40 & 8.92  & 8.96  & 9.68  \\
     & TEMA    & \textbf{16.48} & \textbf{16.39} & \textbf{16.95} & \underline{16.25} & \textbf{14.87} & \textbf{14.05} & \textbf{15.83} \\
     \cmidrule{2-9}
     & CoDA    & \underline{15.62} & \underline{15.80} & \underline{16.64} & \textbf{16.50} & \underline{14.50} & \underline{13.48} & \underline{15.42} \\
    \bottomrule
    \end{tabular}
    \vspace{-2ex}
\end{table}

\begin{table}[t]
    \centering
    \caption{Classification acc. (\%) of different lightweight models on CIFAR10-C using LSQ in quantization.}
    \vspace{-2ex}
    \resizebox{0.95\linewidth}{!}{
    {\fontsize{9pt}{10pt}\selectfont
    \begin{tabular}{l|ccc|ccc}
        \toprule
        \multirow{2}{*}{Method} & \multicolumn{3}{c|}{$k=4$} & \multicolumn{3}{c}{$k=64$} \\
        \cmidrule(lr){2-4} \cmidrule(lr){5-7}
        & 2-bit & 4-bit & 8-bit & 2-bit & 4-bit & 8-bit \\ 
        \midrule
        \multicolumn{7}{l}{\textbf{MobileNet v3-small}} \\
        \midrule
        w/o TTA & 53.38 & 55.46 & 53.89 & 53.38 & 55.46 & 53.89 \\ 
        NORM    & 67.90 & 68.21 & 67.26 & 69.71 & 70.01 & 69.11 \\ 
        TEMA    & 12.13 & 14.63 & 12.14 & 12.31 & 15.31 & 12.31 \\ 
                \midrule
        \ours   & \textbf{74.61} & \textbf{75.42} & \textbf{75.29} &  \textbf{74.49} & \textbf{75.92} & \textbf{75.71} \\ 
        \midrule
        \multicolumn{7}{l}{\textbf{MobileNet v3-large}} \\
        \midrule
        w/o TTA & 56.55 & 55.41 & 56.71 & 56.55 & 55.41 & 56.71 \\ 
        NORM    & 71.21 & 71.37 & 71.72 & 72.86 & 72.96 & 73.27 \\ 
        TEMA    & 10.05 & 16.38 & 16.14 &  9.87 & 16.95 & 16.73 \\
                \midrule
        \ours   & \textbf{76.97} & \textbf{78.96} & \textbf{79.49} & \textbf{77.51} & \textbf{79.40} & \textbf{80.06} \\ 
        \midrule
        \multicolumn{7}{l}{\textbf{EfficientNet-b0}} \\
        \midrule
        w/o TTA & 54.85 & 60.25 & 59.52 & 54.85 & 60.25 & 59.52 \\ 
        NORM    & 69.85 & 72.48 & 72.22 & 71.75 & 74.01 & 73.93 \\ 
        TEMA    & 10.05 & 12.61 & 12.97 &  9.90 & 12.83 & 13.41 \\ 
        \midrule
        \ours   & \textbf{74.00} & \textbf{76.21} & \textbf{76.83} & \textbf{78.59} & \textbf{80.41} & \textbf{80.62} \\ 
        \bottomrule
    \end{tabular}
    }
    }
    \vspace{-3ex}
    \label{tab:results_models}
\end{table}

\subsection{Lightweight and Continual Settings}

\noindent\textbf{Lightweight CNNs.} 
Table~\ref{tab:results_models} evaluates three lightweight CNN architectures that are already designed for resource efficiency, making them a useful test of whether \ours's quantization and adaptation remain beneficial when the backbone is compact. 
\ours achieves the best performance in all tested low-bit settings (window sizes, bit-widths and architectures), verifying its effectiveness under lightweight backbones, while TEMA becomes significantly brittle.

\begin{table}[t]
    \centering
    \caption{Classification acc. (\%) of ResNet18 on ImageNet-C under continually changing target domain, comparing continual TTAs with quantization (LSQ). The model parameters remain unchanged except for stochastic restoration in CoTTA.
    }
    \vspace{-2ex}
    {\fontsize{9pt}{10pt}\selectfont
    \begin{tabular}{l|c|ccc}
        \toprule
        \multirow{1}{*}{Method} &
         32-bit & 2-bit & 4-bit & 8-bit \\
        \midrule
        w/o TTA                         
        & 22.36 & 16.25 & 19.90 & 20.15 \\
        NORM
        & 29.13 & 24.44 & 28.50 & 29.23 \\
        TEMA & 31.67 & 28.85 & 31.89 & 32.34 \\
        TENT
        & 32.76 & - & - & - \\
        CoTTA
        & \textbf{34.14} & - & - & - \\
        \midrule
        \ours & 33.79 & \textbf{31.35} & \textbf{33.43} & \textbf{34.07} \\
        \bottomrule
    \end{tabular}
    }
    \vspace{-3ex}
    \label{tab:results_continual}
\end{table}

\vspace{1ex}\noindent\textbf{Continual Shifts.} 
Table~\ref{tab:results_continual} evaluates continual ImageNet-C shifts, where target domains arrive sequentially without resetting the model between domains. Following CoTTA~\cite{wang2022continual}, we evaluate the continual shifts in the order of Gaussian noise $\rightarrow$ shot noise $\rightarrow$ impulse noise $\rightarrow$ defocus blur $\rightarrow$ glass blur $\rightarrow$ motion blur $\rightarrow$ zoom blur $\rightarrow$ snow $\rightarrow$ frost $\rightarrow$ fog $\rightarrow$ brightness $\rightarrow$ contrast $\rightarrow$ elastic transform $\rightarrow$ pixelate $\rightarrow$ JPEG compression {at severity level 3 (50,000 images per corruption). Each batch window contains samples from a single corruption domain; at domain boundaries, the adaptation statistics are carried forward without reset.}

\ours is not designed specifically for catastrophic-forgetting mitigation but it performs the strongest in 2/4/8-bit continual settings. 
Furthermore, \ours with 4-bit quantization performs better than TENT, and \ours with 8-bit quantization performs comparably to CoTTA. Note that both TENT and CoTTA require FP models and backpropagation.  
This supports the claim that separating source-stable LFC statistics from target-specific HFC statistics is useful not only for episodic shifts but also for sequential domain changes.

\begin{table}[t]
    \centering
    \caption{Latency of ResNet18/50 under various TTA methods with PyTorch quantization. BP refers to back-propagation. (RPi: Raspberry Pi 4B, $k=16$)}
    \vspace{-2ex}
    \resizebox{\linewidth}{!}{
    \begin{tabular}{ccccccc}
    \toprule
    \textbf{Model} & \textbf{Method} & \textbf{BP}& \textbf{Acc. (\%)} & \textbf{RPi (s)} & \textbf{Jetson (s)} & \textbf{Laptop (s)} \\ 
    \midrule
    \multirow{7}{*}{\makecell{ResNet18 \\ (32-bit)}}
     & w/o TTA  &               & 22.36 & 6.81  & 5.21 & 0.54  \\      
     \cmidrule{2-7}
     & NORM     &               & 29.13 & 7.35  & 5.51  & 0.96  \\ 
     & TEMA     &               & 29.33 & 20.53 & 12.72  & 1.60  \\ 
     & TENT     & \checkmark    & 30.62 & 15.45 & 6.42  & 1.19  \\ 
     & SAR      & \checkmark    & 33.88 & 31.12 & 12.94 & 2.33  \\ 
     \cmidrule{2-7}
     % & CoDA* & & 33.87 & - & 2.25 \\ 
     & \ours    &               & 33.80 & 11.24 & 6.54  & 1.40  \\ 
     \midrule
     \multirow{5}{*}{\makecell{ResNet18 \\ (8-bit)}}
     & w/o TTA  &               & 20.15 & 3.33  & 2.03  & 0.27  \\ 
     \cmidrule{2-7}
     & NORM     &               & 29.23 & 4.18  & 2.27  & 0.70  \\ 
     & TEMA     &               & 25.36 & 12.83 & 5.86 & 1.02  \\ 
     \cmidrule{2-7}
     % & CoDA* & & 34.19 & - & 1.92 \\ 
     & \ours    &               & 34.17 & 8.12 & 3.53  & 1.16  \\ 
    \midrule
    \multirow{7}{*}{\makecell{ResNet50 \\ (32-bit)}}
     & w/o TTA  &               & 27.74 & 17.63 & 19.21 & 1.31  \\ 
     \cmidrule{2-7}
     & NORM     &               & 36.20 & 19.41 & 20.32 & 3.23  \\ 
     & TEMA     &               & 34.96 & 87.05 & 51.62 & 9.72  \\ 
     & TENT     & \checkmark    & 38.55 & 40.01 & 22.54 & 3.44  \\ 
     & SAR      & \checkmark    & 41.49 & 81.01 & 45.32 & 6.85  \\ 
     \cmidrule{2-7}
     % & CoDA* & & 40.76 & 33.91 & 8.46 \\ 
     & \ours    &               & 40.73 & 23.96 & 21.33 & 3.83  \\ 
     \midrule
     \multirow{5}{*}{\makecell{ResNet50 \\ (8-bit)}}
     & w/o TTA  &               & 22.58 & 8.55 & 4.65  & 0.65  \\ 
     \cmidrule{2-7}
     & NORM     &               & 31.87 & 12.76 & 5.84  & 2.70  \\ 
     & TEMA     &               & 26.83 & 67.71 & 20.98 & 7.83  \\ 
     \cmidrule{2-7}
     % & CoDA* & & 36.65 & - & 7.92 \\ 
     & \ours    &               & 36.68 & 16.58 & 7.06  & 3.37  \\ 
    \bottomrule
    \end{tabular}
    }
    \label{tab:latency}
% }
\vspace{-1ex}
\end{table}

%%%%%%%%%% System 
\section{Experiments: Systems Performance}\label{sec:system}

We now evaluate the system cost of the same adaptation methods under the deployment protocol in Section~\ref{sec:implementation}. We measure wall-clock latency, peak memory, and CPU utilization on Raspberry Pi 4B, Jetson Orin Nano CPU, and a laptop CPU reference platform. All measurements include both inference and adaptation: \ours includes the first-layer frequency split, FABN statistic update, and remaining quantized forward pass; forward-only baselines include their statistic updates; and backpropagation-based baselines include backward computation and optimizer updates.

\begin{figure}[t]
    \centering
    \includegraphics[width=\linewidth]{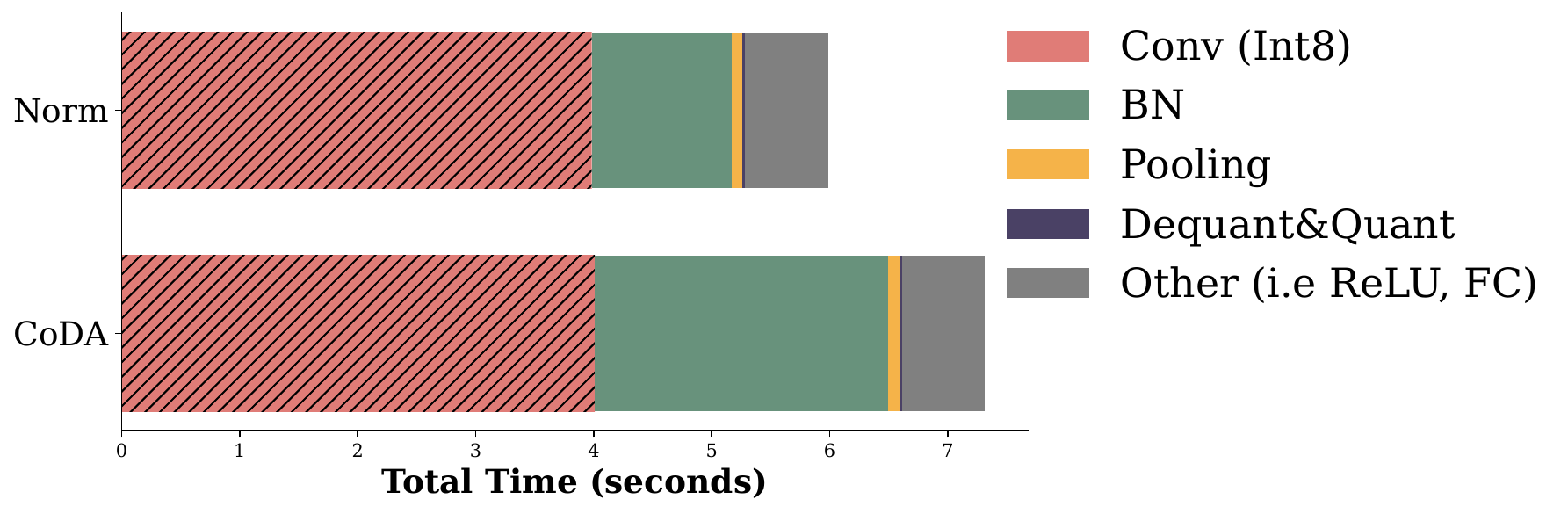}
    \vspace{-3ex}
    \caption{Runtime breakdown of NORM and \ours with PyTorch quantization on Jetson Orin Nano (ResNet50, $k$=16).}
    \vspace{-2ex}
    \Description{Horizontal stacked bar chart of total runtime in seconds on Jetson Orin Nano for NORM and CoDA, split into INT8 convolution, batch normalization, pooling, dequantize and quantize, and other operations. Both bars spend about 4 seconds in INT8 convolution. CoDA totals about 7 seconds versus about 6 seconds for NORM, with the extra time coming mainly from a longer batch normalization segment.}
    \label{fig:breakdown}
\end{figure}

%%%%%%%%%%%%%%%%%%%%%%%%%%%%%%%%%%%%%%%
%%%%%%%%%%%%%%%%%%%%%%%%%%%%%%%%%%%%%%%
\subsection{Latency--Memory Trade-off} 

Table~\ref{tab:latency} reports end-to-end latency at \(k=16\). NORM is the minimum-latency method because it performs only standard BN-statistic adaptation. \ours is not intended to be faster than NORM; instead, it trades a bounded first-layer FABN overhead for substantially higher robustness while preserving fixed quantized weights and forward-only adaptation. 
Across Raspberry Pi 4B and Jetson Orin Nano, \ours remains much faster than TEMA and backpropagation-based upper bounds in the main ResNet18/50 settings. For example, on Jetson Orin Nano with ResNet50 8-bit, \ours reaches 36.68\% accuracy with 7.06s latency, compared with 31.87\% and 5.84s for NORM, 26.83\% and 20.98s for TEMA, and substantially higher latency for backpropagation-based methods in full precision. This illustrates \ours's intended operating point: higher accuracy than lightweight forward-only normalization baselines, while avoiding the training-time machinery required by TENT and SAR.

Figure~\ref{fig:breakdown} breaks down \ours's runtime on Jetson Orin Nano. The additional cost over NORM is concentrated in the BN/FABN portion, where \ours performs the first-layer frequency split and statistic update. Other components remain close to the standard quantized inference path. This supports the design choice in Section~\ref{sec:method}: frequency-aware processing is restricted to the first BN layer, while later layers use lightweight statistic updates.

\begin{figure}[t]
    \centering
    \includegraphics[width=\linewidth]{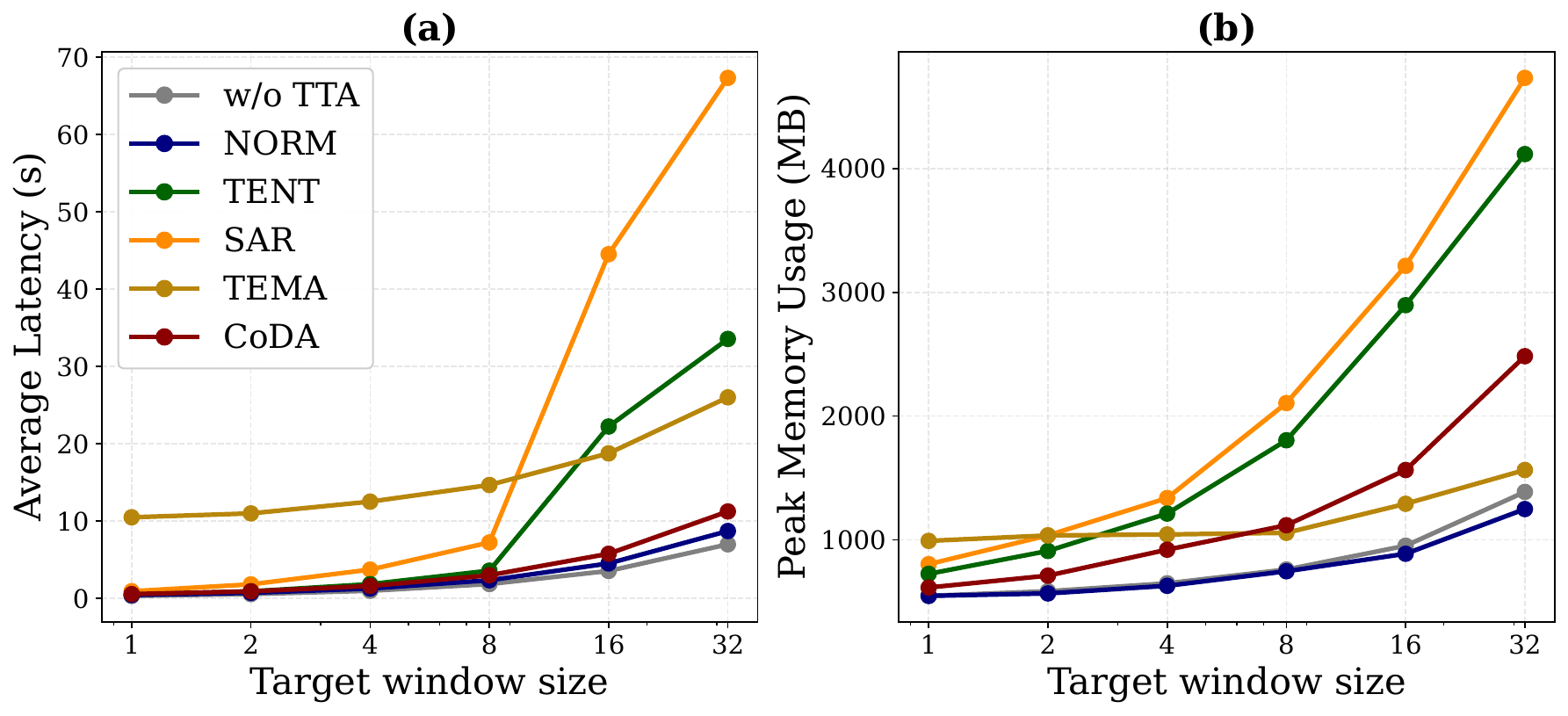}
    \vspace{-5ex}
    \caption{Latency and peak memory of various TTA methods using ResNet50 under different $k$. }
    \vspace{-4ex}
    \Description{Two line plots versus target window size from 1 to 32 for ResNet50 on an edge device. Left, average latency: SAR rises sharply to about 67 seconds at size 32, TENT to about 33 seconds, and TEMA to about 26 seconds, while CoDA stays near NORM and no adaptation, at about 11 seconds or less. Right, peak memory: SAR and TENT grow to about 4.7 and 4.1 gigabytes at size 32, CoDA reaches about 2.5 gigabytes, and NORM, TEMA, and no adaptation remain below about 1.6 gigabytes.}
    \label{fig:latency_peakmem}
\end{figure}

\begin{figure}[t]
    \centering
    \begin{subfigure}{0.9\linewidth}
        \centering
        \includegraphics[width=\linewidth]{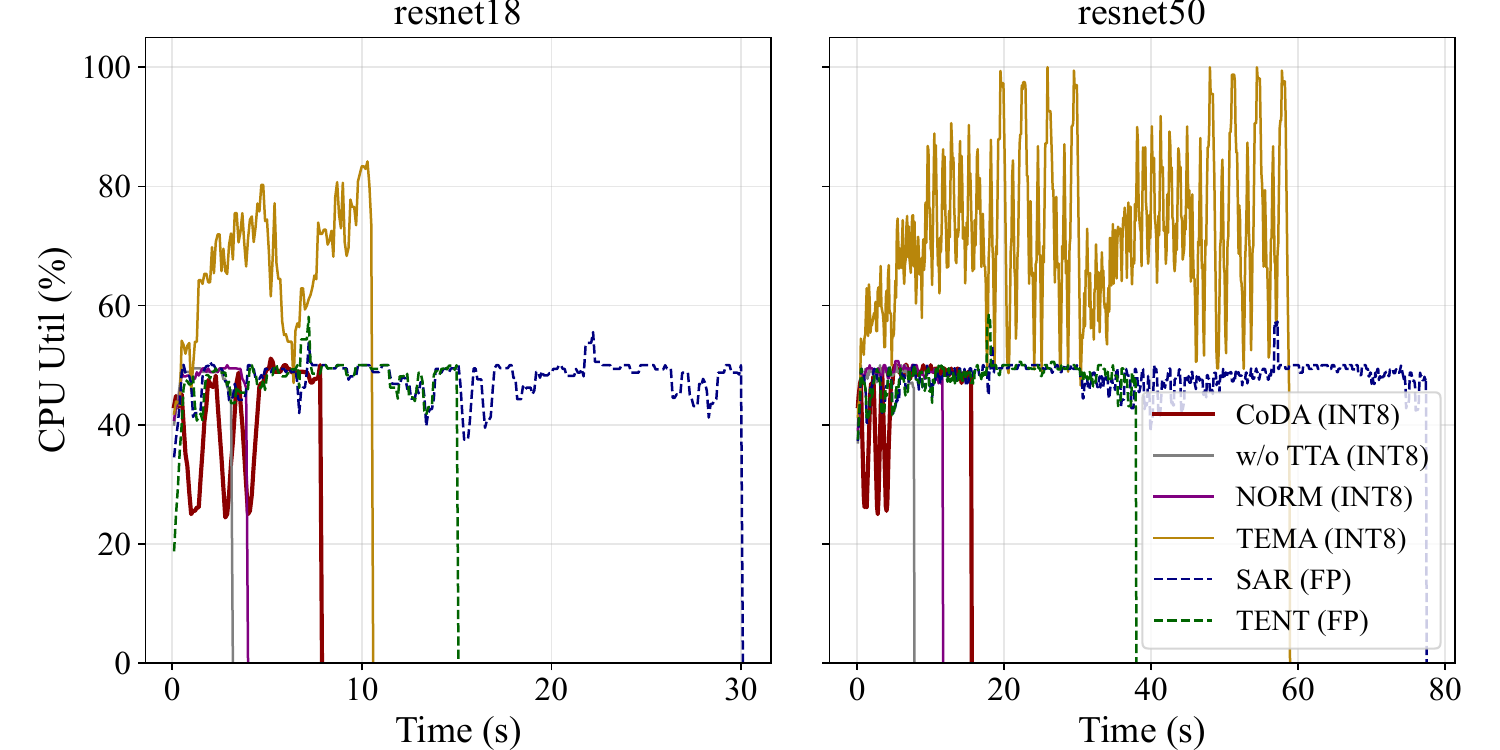}
        \caption{CPU Utilization on Raspberry Pi 4B.}
        \label{fig:cpu_util_rpi}
    \end{subfigure}
    \begin{subfigure}{0.9\linewidth}
        \centering
        \includegraphics[width=\linewidth]{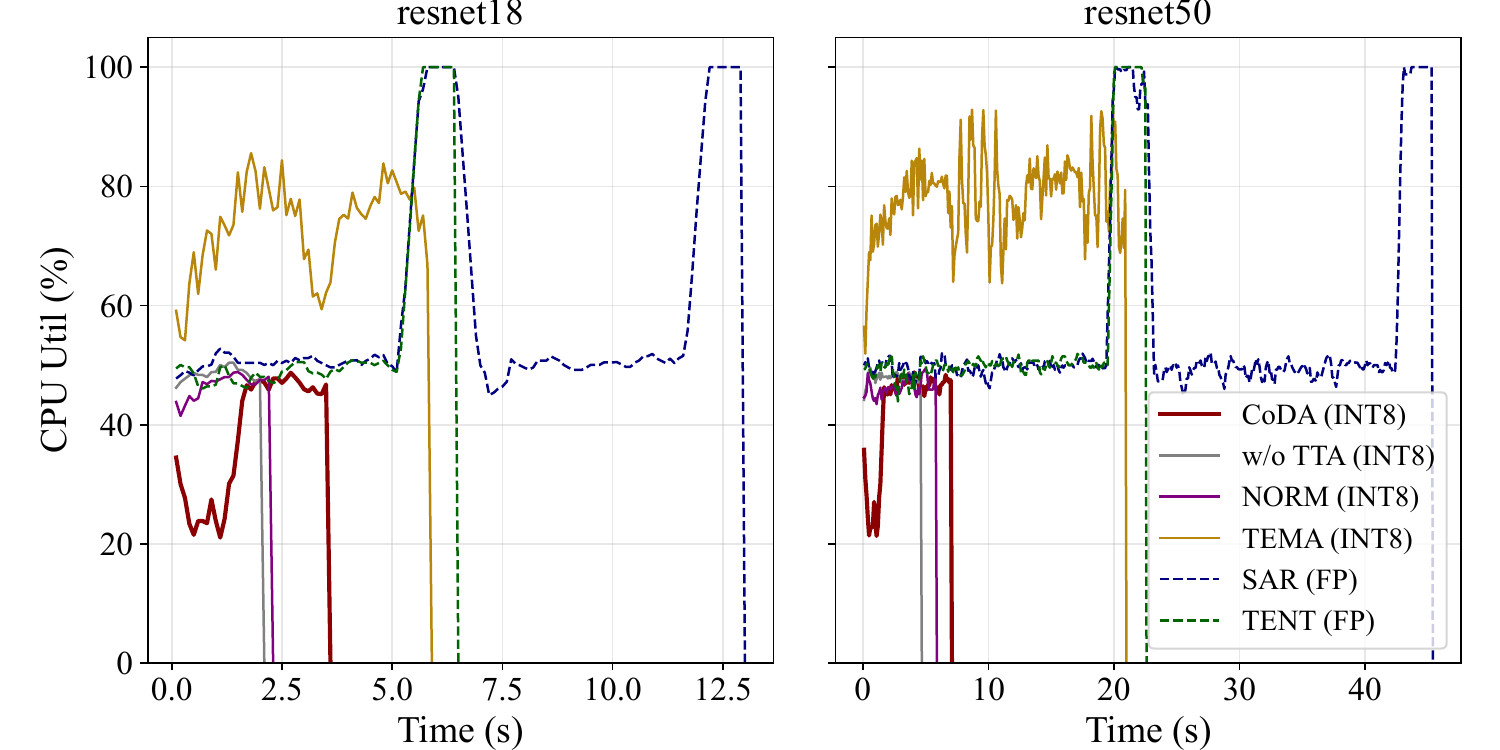}
        \caption{CPU Utilization on Jetson Orin Nano.}
        \label{fig:cpu_util_jetson}
    \end{subfigure}
    \vspace{-2ex}
    \Description{Two rows of CPU utilization traces over time, Raspberry Pi 4B on top and Jetson Orin Nano below, each with ResNet18 and ResNet50 panels. INT8 methods, including CoDA, NORM, and no adaptation, finish first and stay at or below about 50 percent utilization. TEMA runs at higher and noisier utilization, up to 80 to 100 percent. The full-precision methods SAR and TENT run the longest, up to roughly 80 seconds on Raspberry Pi and 45 seconds on Jetson with ResNet50.}
    \caption{CPU utilization on the edge device during a single inference/adaptation run when $k$=16.} 
    \label{fig:cpu_util}
    \vspace{-4ex}
\end{figure}

Figure~\ref{fig:latency_peakmem} evaluates latency and peak memory as the adaptation window \(k\) changes. \ours's latency grows moderately with \(k\) and remains close to NORM, while TEMA has a fixed two-forward-pass overhead even at small \(k\). \ours's peak memory increases with \(k\) because FABN stores first-layer frequency features for the current target window. This overhead is bounded by the first activation feature map and does not require gradients, optimizer state, or mutable weights. In small-window settings, which are common in on-device TTA~\cite{danilowski2026botta}, \ours remains more memory-efficient than TEMA while providing stronger accuracy than NORM.

%%%%%%%%%%%%%%%%%%%%%%%%%%%%%
%%%%%%%%%%%%%%%%%%%%%%%%%%%%%
\subsection{CPU Utilization} %  CPU utilization

Figure~\ref{fig:cpu_util} shows CPU utilization during a single inference/adaptation run on Raspberry Pi 4B and Jetson Orin Nano. \ours follows a utilization pattern similar to NORM for most of the forward pass, with an additional phase at the beginning for first-layer FABN. TEMA uses only forward passes but keeps the CPU active for longer because it requires two passes. TENT and SAR incur additional compute from backpropagation and optimizer updates. These measurements confirm that \ours exposes adaptation through a bounded forward-only statistic update rather than through online training.

\vspace{1ex}\noindent\textbf{Takeaway.} 
\ours is not the lowest-latency adaptation method; NORM is. \ours's advantage is the deployment-compatible accuracy--system-cost trade-off: it improves robustness over forward-only normalization baselines while keeping convolution and linear layers quantized, avoiding labels, source data, gradients, optimizer state, and weight updates. The overhead is concentrated in the first-layer FABN and remains bounded under small target windows.
\section{Analysis and Ablations}
\label{sec:analysis}

This section explains why \ours's main design choices matter. We focus on four questions: (1) why source-time QAT uses low-frequency reconstructions, (2) why FABN uses both low- and high-frequency target statistics, (3) whether LFC-QAT and FABN each contribute to the final result, and (4) why \ours applies frequency-aware adaptation only at the first BN layer.

\begin{table}[t]
    \centering
    \caption{Frequency analysis showing classification acc. (\%) of 2-bit models trained with varying $r$ for LFCs and HFCs.
    }
    \vspace{-2ex}
    {\fontsize{9pt}{10pt}\selectfont
    \setlength{\tabcolsep}{1mm}  % 기본값은 6pt 정도
    \begin{tabular}{lccccccccc}
        \toprule
        \multirow{2}{*}{\makecell{Filter\\type}} & \multicolumn{4}{c}{CIFAR10 | ResNet26} & \multicolumn{4}{c}{ImageNet | ResNet50} \\
        \cmidrule(lr){2-5} \cmidrule(lr){6-9}
        & $r$ & FP & LSQ & LQ & $r$ & FP & LSQ & LQ \\
        \midrule
        No filter & - & 90.35 & 86.11 & 89.06 & - & 66.96 & 60.14 & 63.35 \\
        \midrule
        \multirow{4}{*}{\makecell{Low\\pass\\filter}} 
        & 4  & 65.66 & 68.99 & 71.94 & 28  & 59.26 & 50.61 & 55.51 \\
        & 8  & 81.98 & 80.82 & 84.73 & 56  & 64.91 & 58.55 & 61.83 \\
        & 12 & 85.07 & 83.95 & 87.48 & 84  & 66.24 & 60.05 & 62.73 \\
        & 16 & 86.83 & 83.46 & 88.94 & 112 & 66.93 & 60.51 & 63.63 \\
        \midrule
        \multirow{4}{*}{\makecell{High\\pass\\filter}} 
        & 4  & 74.07 & 54.35 & 66.45 & 28  & 0.17 & 0.20 & 0.52 \\
        & 8  & 21.21 & 12.76 & 19.72 & 56  & 0.11 & 0.13 & 0.23 \\
        & 12 & 11.02 & 12.00 & 11.96 & 84  & 0.15 & 0.11 & 0.22 \\
        & 16 & 10.11 & 10.94 & 9.63  & 112 & 0.15 & 0.13 & 0.15 \\
        \bottomrule
    \end{tabular}%
    }
    \vspace{-3.5ex}
    \label{tab:frequency_analysis}
\end{table}

%%%%%%%%%%%%%%%%%%%%%%%%%%%%%%%%%%
%%%%%%%%%%%%%%%%%%%%%%%%%%%%%%%%%%
\subsection{Why Low-Frequency Source Training?}

\ours trains the quantized model on low-frequency source reconstructions because low-bit models have limited representational capacity. The goal is to allocate this capacity to domain-stable visual structure rather than source-specific high-frequency details.

Table~\ref{tab:frequency_analysis} compares full-precision (FP) and 2-bit QAT models trained on LFCs and HFCs under different radius values \(r\), then evaluated on full-frequency clean test data. HFC-only training is unreliable for low-bit models. On ImageNet, HFC-only accuracy is near zero across radii and quantization methods. On CIFAR10, HFCs contain some task signal at small radius, especially in full precision, but LSQ and LQ degrade substantially compared with LFC training. 
In contrast, LFC training preserves much stronger clean accuracy across both CIFAR10 and ImageNet. These results support using LFCs, rather than HFCs, as the source-time training signal for low-bit models.

\begin{table}[t]
    \centering
    {\fontsize{9pt}{10pt}\selectfont
    \setlength{\tabcolsep}{1mm}  % 기본값은 6pt 정도
    \caption{
    Frequency analysis showing classification acc. (\%) of FP and 2-bit models trained with varying $r$ for LFCs.
    }
    \vspace{-2ex}
    \begin{tabular}{lccccccccc}
        \toprule
        \multirow{2}{*}{\makecell{Filter\\type}} & \multicolumn{4}{c}{CIFAR10-C | ResNet26} & \multicolumn{4}{c}{ImageNet-C | ResNet50} \\
        \cmidrule(lr){2-5} \cmidrule(lr){6-9}
        & $r$ & FP & LSQ & LQ & $r$ & FP & LSQ & LQ \\
        \midrule
        No filter & - & 53.86 & 50.15 & 54.35 & - & 27.74 & 19.76 & 22.23 \\
        \midrule
        \multirow{4}{*}{\makecell{Low\\pass\\filter}} 
        & 4  & 55.24 & 59.51 & 60.60 & 28  & 30.40 & 21.90 & 25.23 \\
        & 8  & 64.61 & 65.45 & 67.53 & 56  & 29.76 & 20.82 & 25.06 \\
        & 12 & 63.42 & 62.24 & 63.87 & 84  & 28.06 & 19.77 & 23.20 \\
        & 16 & 56.77 & 55.20 & 58.10 & 112 & 28.08 & 19.72 & 23.38 \\
        \bottomrule
    \end{tabular}
    \label{tab:frequency_analysis_c}
    }    
    \vspace{-2ex}
\end{table}

Table~\ref{tab:frequency_analysis_c} evaluates the same models under corruption shifts. LFC training gives large robustness gains on CIFAR10-C and modest but useful gains in the main ImageNet-C low-bit settings. This suggests that low-frequency source training is not simply removing information; rather, it biases the quantized model toward structure that \textit{transfers better} under target-domain shifts.

\begin{figure}[t]
  \centering
  \begin{subfigure}{.48\linewidth}
  \includegraphics[width=\linewidth]{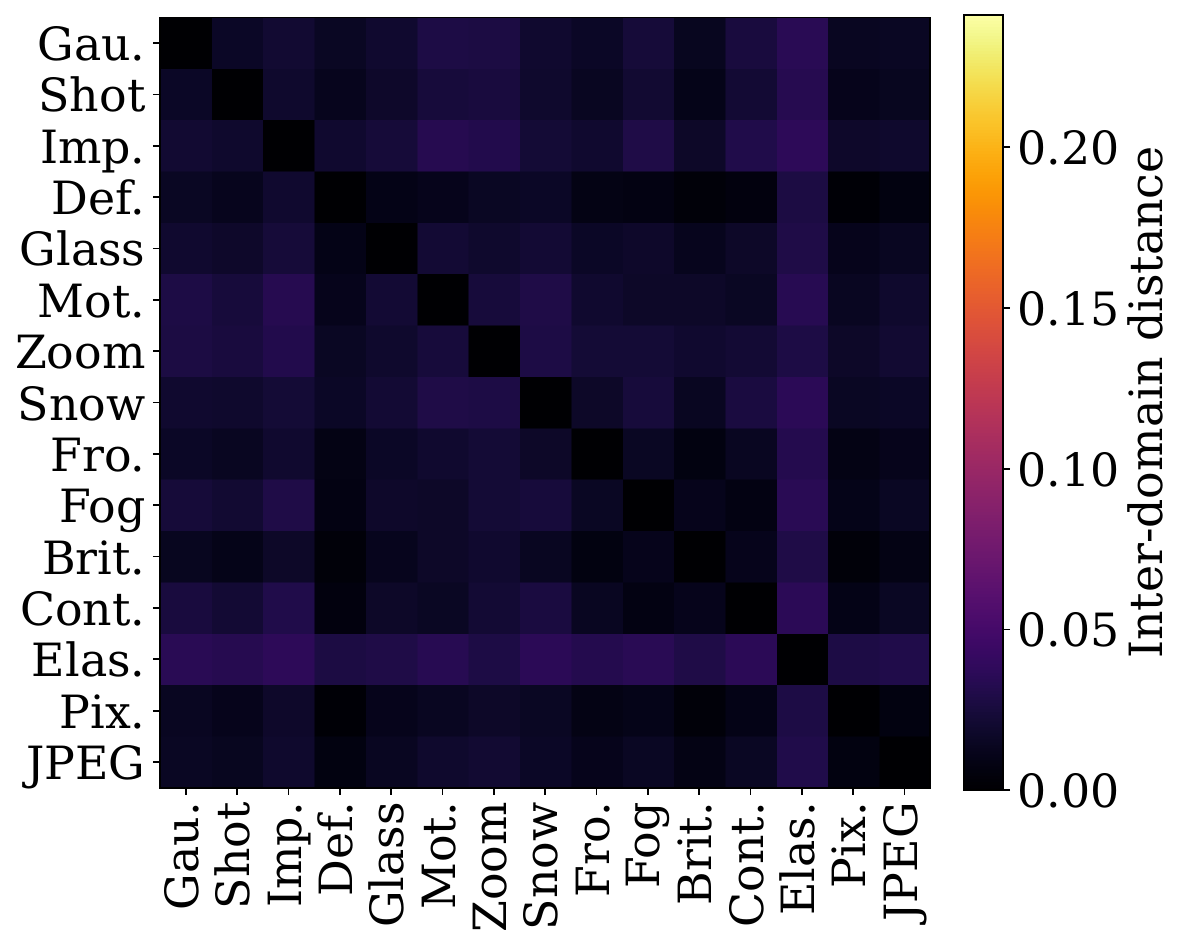}
  \caption{LFC distances.}
  \label{fig:low_pass_distance}
  \end{subfigure}
  \hspace{0ex}
  \begin{subfigure}{.48\linewidth}
  \includegraphics[width=\linewidth]{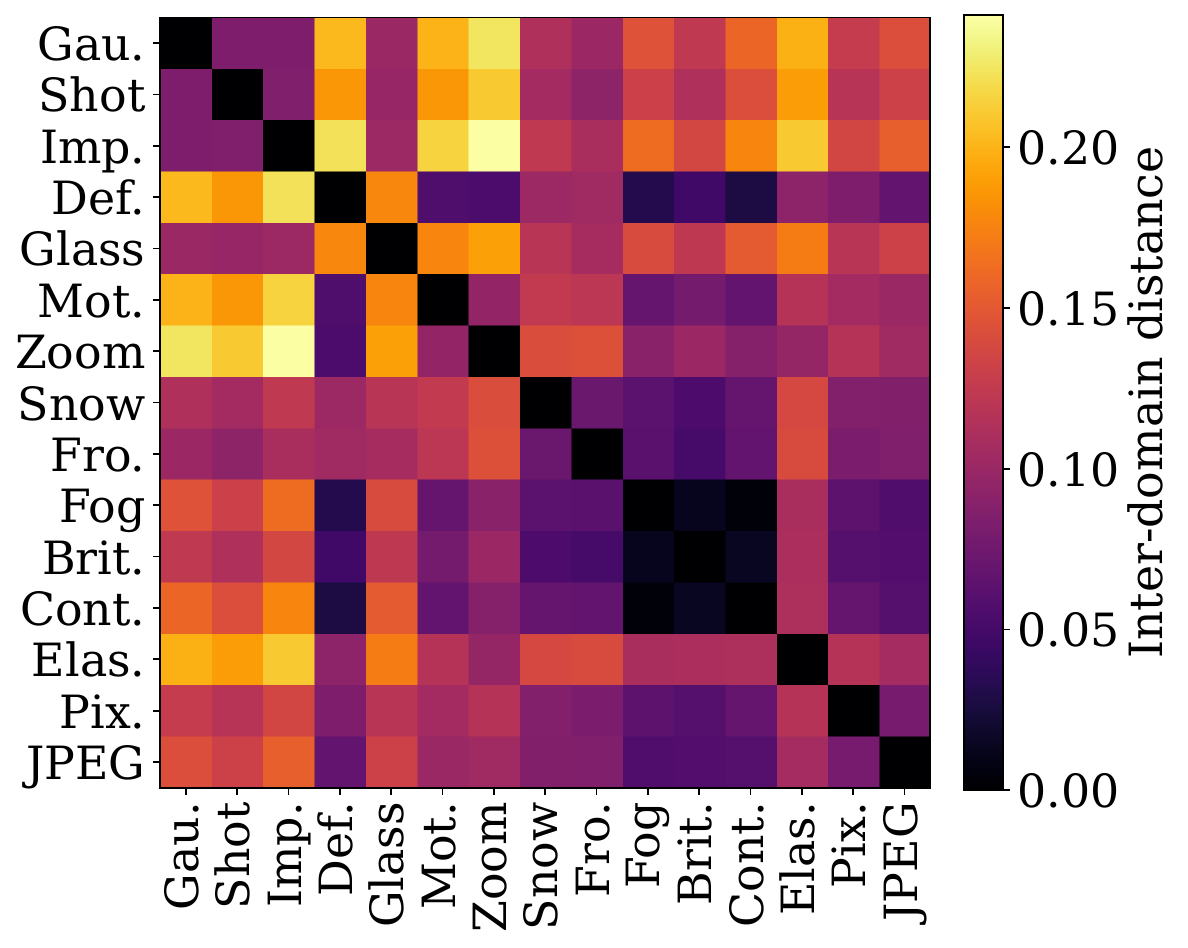}
  \caption{HFC distances.}
  \label{fig:high_pass_distance}
  \end{subfigure}
  \vspace{-2ex}
  \caption{Inter-domain distance matrices of LFCs and HFCs in the frequency domain. 
  }
  % \vspace{-3ex}
  \Description{Two 14 by 14 heatmaps of inter-domain distances among CIFAR10-C corruption types, on a shared color scale from 0 to about 0.24. The left heatmap, for low-frequency components, is almost uniformly dark, meaning distances are near zero across all corruptions. The right heatmap, for high-frequency components, has many bright cells, especially between the noise corruptions and the blur or elastic corruptions, meaning high-frequency components differ much more across domains.}
  \label{fig:distance_matrix}
\end{figure}

\begin{figure}[t]
  \centering
  \begin{subfigure}{.24\linewidth}
  \includegraphics[width=\linewidth]{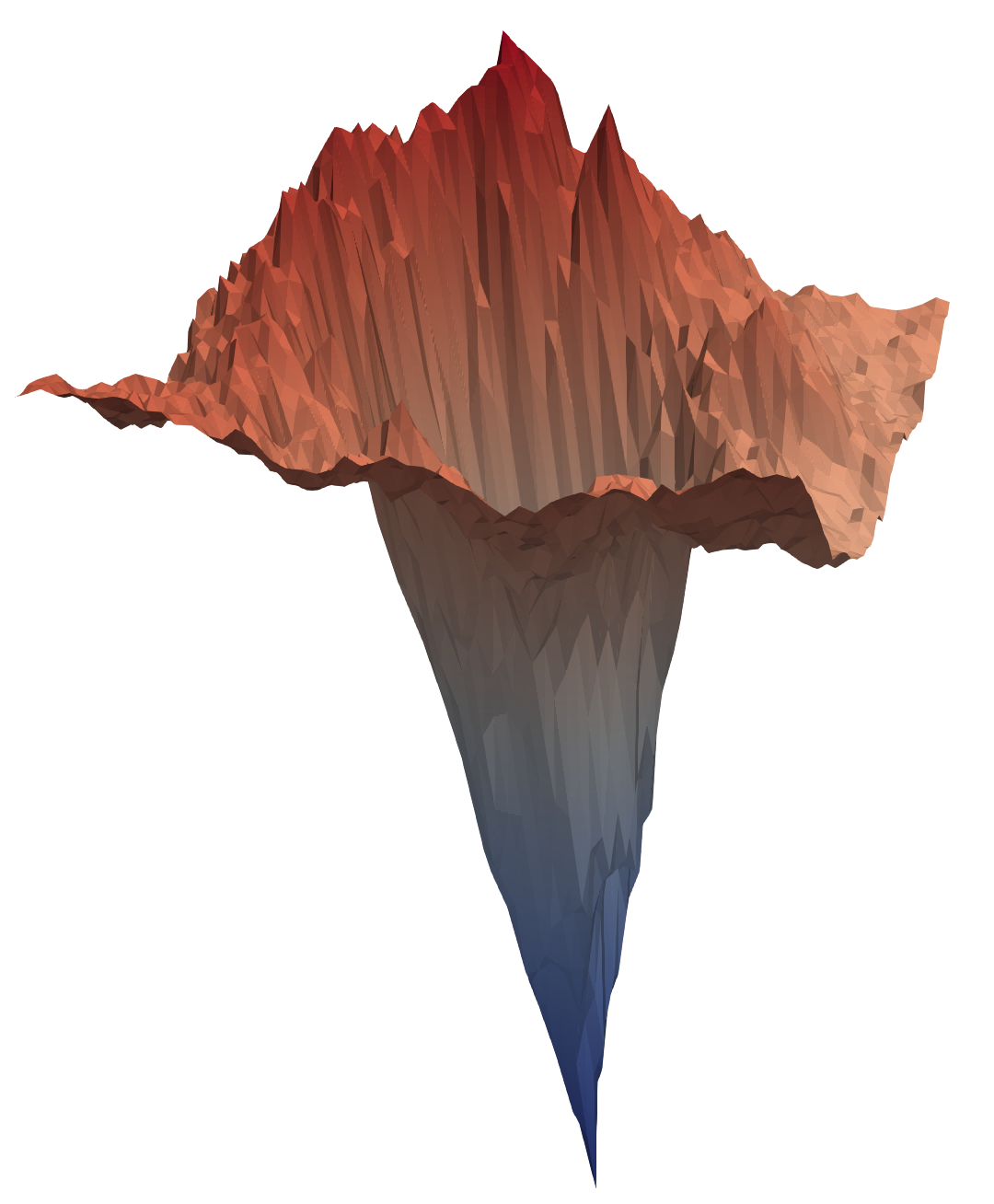}
  \caption{FFC Clean.}
  \label{fig:corruptiona}
  \end{subfigure}
  \begin{subfigure}{.24\linewidth}
  \includegraphics[width=\linewidth]{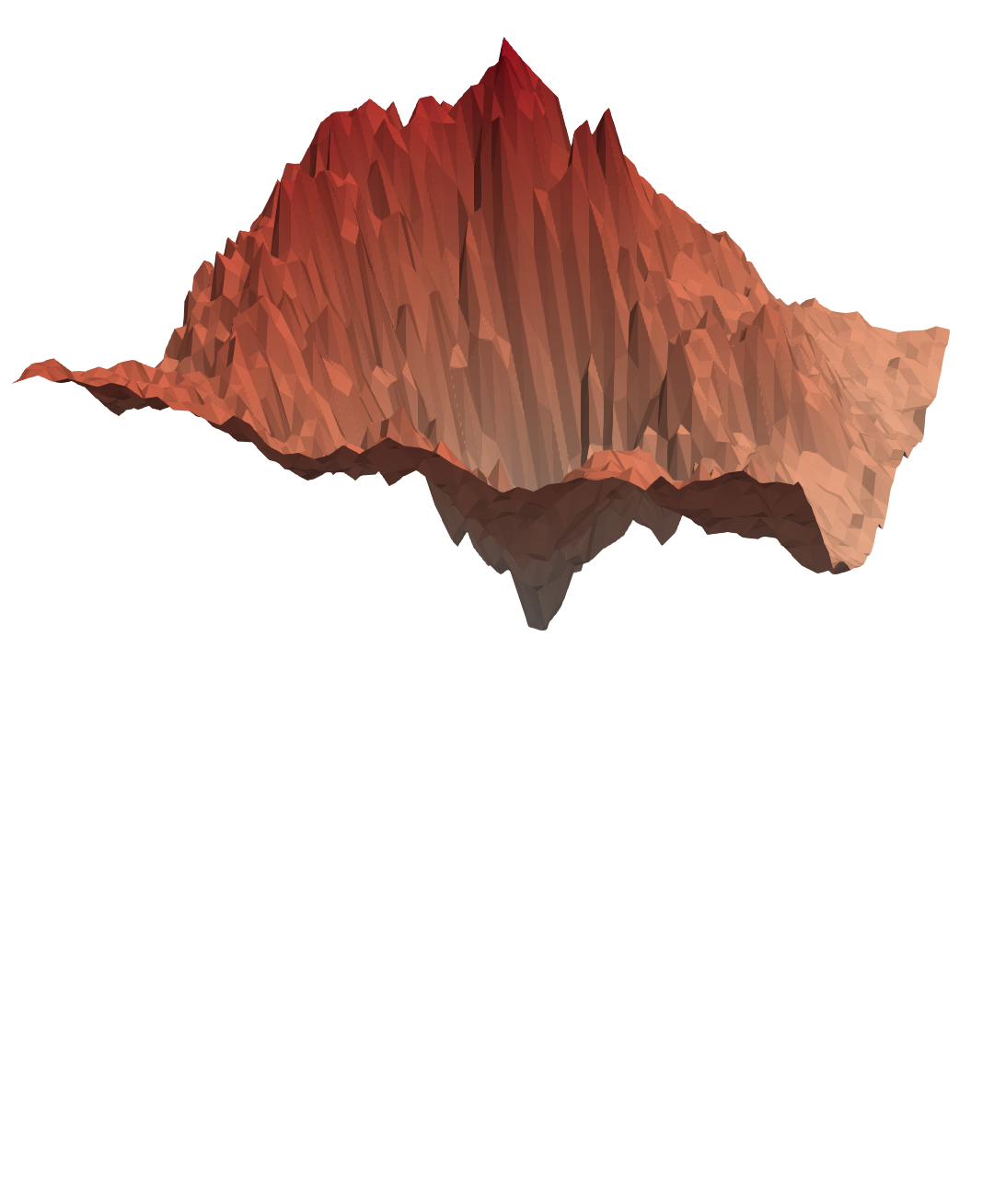}
  \caption{FFC Corr.}
  \label{fig:corruptionb}
  \end{subfigure}
  \begin{subfigure}{.24\linewidth}
  \includegraphics[width=\linewidth]{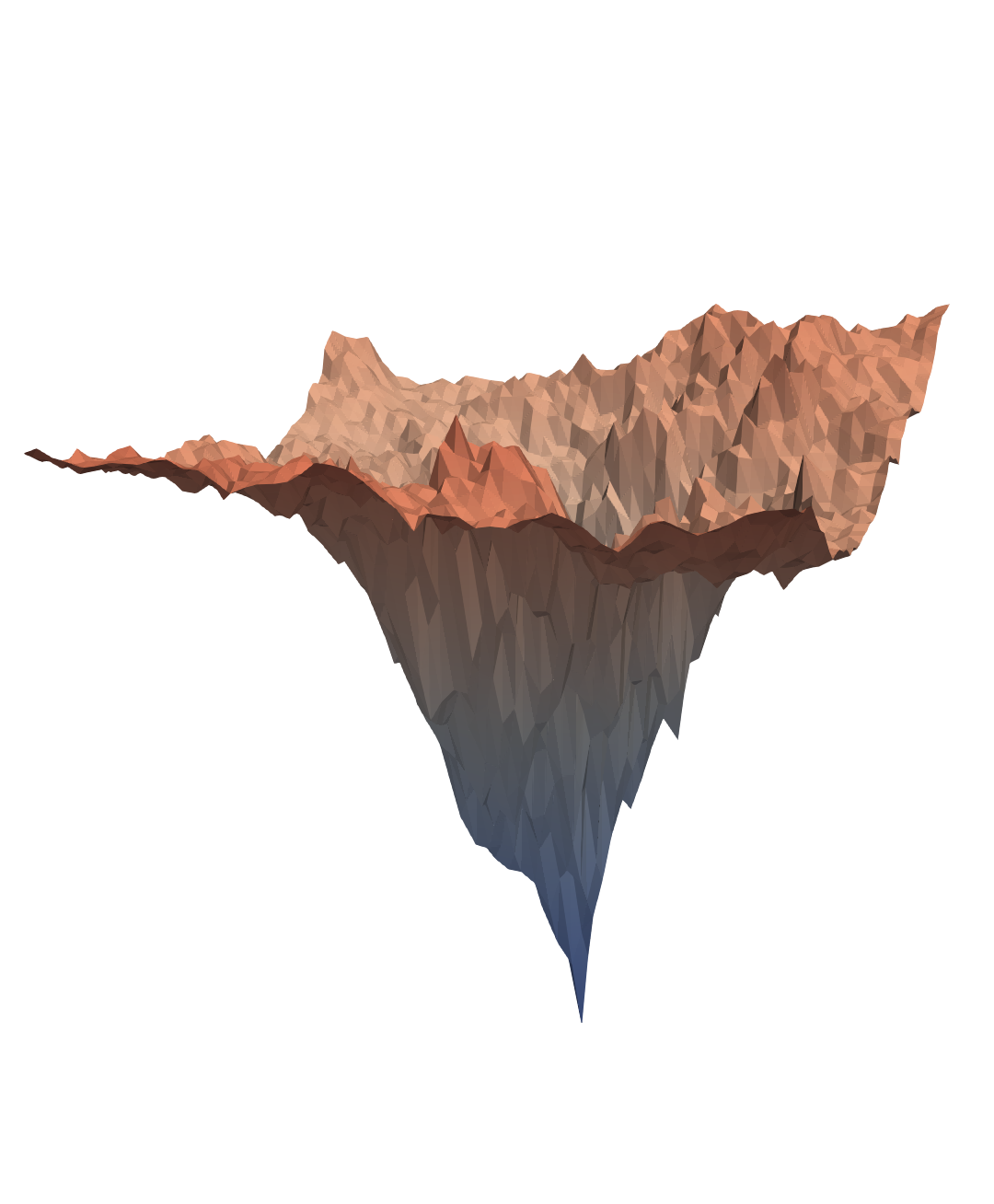}
  \caption{LFC Clean.}
  \label{fig:corruptionc}
  \end{subfigure}
  \begin{subfigure}{.24\linewidth}
  \includegraphics[width=\linewidth]{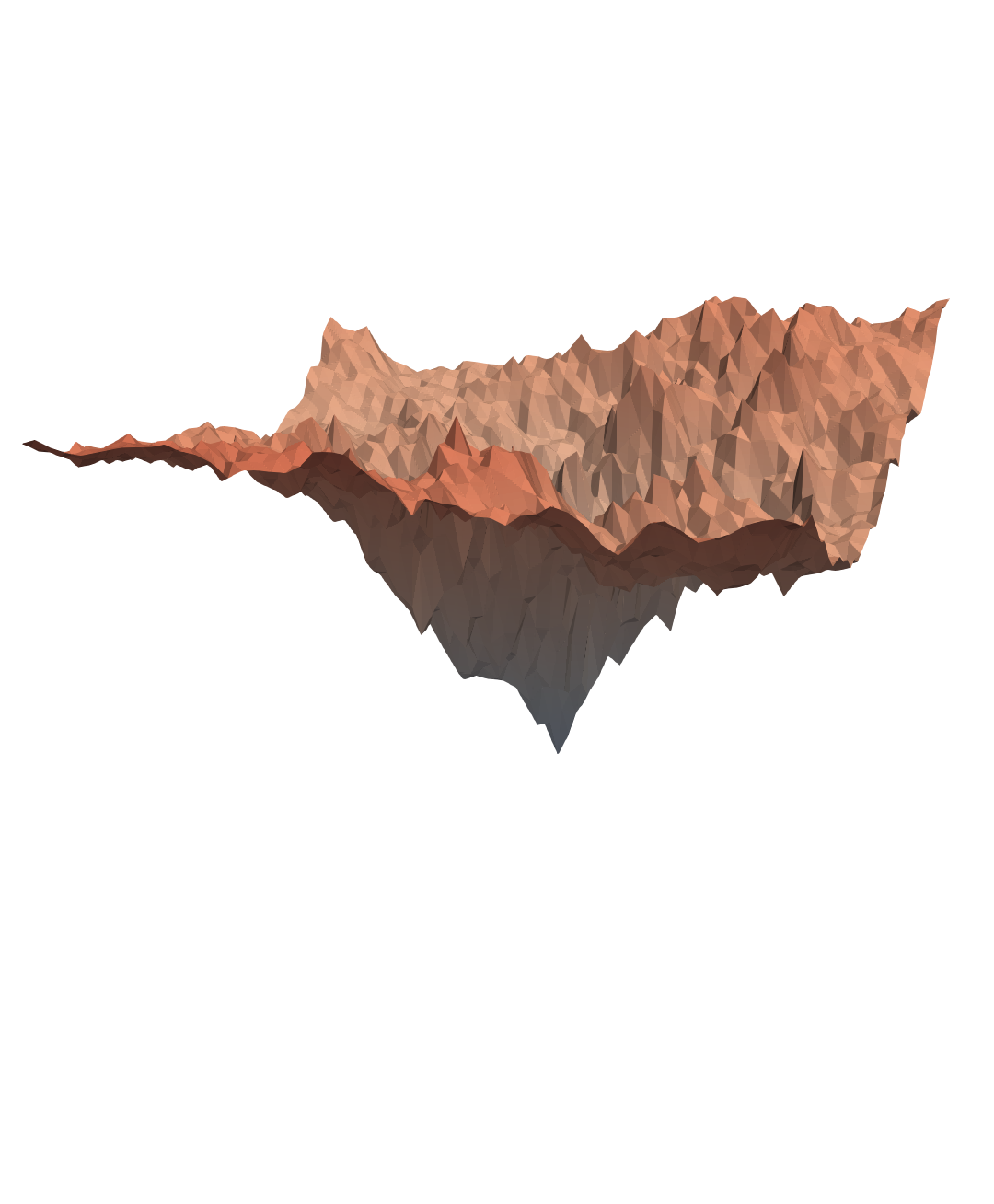}
  \caption{LFC Corr.}
  \label{fig:corruptiond}
  \end{subfigure}
  \vspace{-2ex}
  \Description{Four 3D loss-landscape surface plots of a 2-bit ResNet26 trained with LSQ: full-frequency training on clean and on corrupted test data, and low-frequency training on clean and on corrupted test data. Each surface shows a funnel-shaped basin around the trained weights; the plots compare how the shape and sharpness of the basin change between clean and corrupted data for the two training frequencies.}
  \caption{
  The loss landscapes of 2-bit ResNet26 (with LSQ QAT) on CIFAR10 (Clean) and CIFAR10-C (Corruption) test data with different training frequencies. $r$ for LFCs is 8.
  }
  \label{fig:loss}
  \vspace{-3ex}
\end{figure}

Figures~\ref{fig:distance_matrix} and~\ref{fig:loss} provide additional evidence. 
In Figure~\ref{fig:distance_matrix}, LFCs show smaller inter-domain distances—measured as the cosine distance in the Fourier-transformed frequency domain—across corruptions than HFCs, indicating that low-frequency components vary less across target corruptions. 
Figure~\ref{fig:loss} illustrates the robustness of LFC-QAT via loss landscape visualization~\cite{li2018visualizing}.
On clean images (Figures~\ref{fig:corruptiona} and~\ref{fig:corruptionc}), the FFC-trained model exhibits a sharper, partially concave landscape, suggesting it may quickly lose its perforamance under domain shifts.
On corrupted images (Figures~\ref{fig:corruptionb} and~\ref{fig:corruptiond}), the FFC-trained model shows a deteriorated minimum, indicating it must substantially reshape its landscape to handle domain shifts, whereas the LFC-trained model's smoother, more convex landscape reflects a more malleable state well-prepared for future adaptation.
% In Figure~\ref{fig:loss}, the loss-landscape comparison further shows that LFC-trained 2-bit models behave more smoothly under corrupted inputs than FFC-trained models. 
Together, these results support \ours's source-time design: LFC-QAT prepares a compressed representation that is more stable under deployment-time shifts.

% visualizations

%%%%%%%%%%%%%%%%%%%%%%%%%
%%%%%%%%%%%%%%%%%%%%%%%%%
\subsection{Why Frequency-Aware Target Adaptation?}

LFC-QAT improves the source-time representation, but target-domain sensing conditions still change after deployment. FABN therefore adapts BN statistics using target-window information. The key question is whether FABN should use only LFCs, only HFCs, or the full target activation.

Table~\ref{tab:results_filters} varies the frequency content used by FABN at test time for a 2-bit ResNet26 trained with LFC-QAT on CIFAR10. 
Using the full target feature gives the best adaptation accuracy, while LFC-only or HFC-only adaptation is weaker. In particular, when the test-time LFC radius is smaller than the training radius ($r$=8), adaptation degrades, and HFC-only adaptation performs poorly.
This supports the frequency-asymmetric design: source-time training should suppress unreliable source HFCs, but deployment-time adaptation should not discard target HFCs. LFC target statistics preserve source-trained structure, while HFC target statistics capture current sensing conditions such as noise, blur, texture, and compression artifacts.

\begin{figure}[t]
    \centering
    \includegraphics[width=\linewidth]{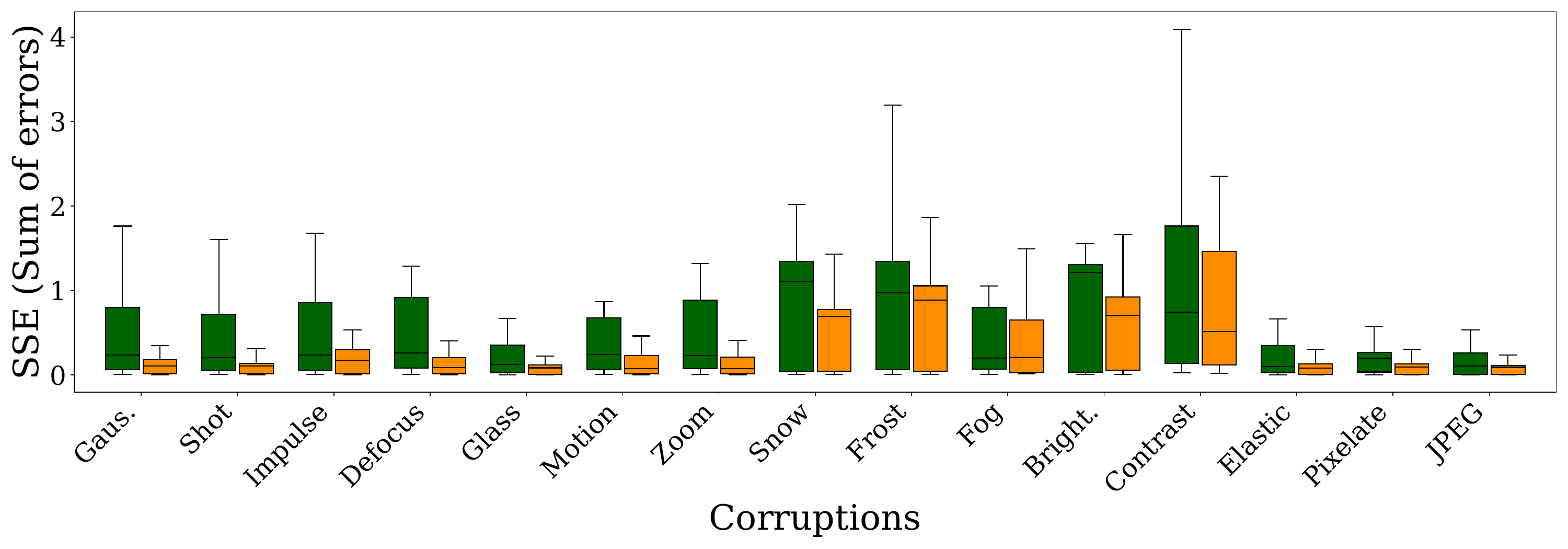}    % \centering
    \vspace{-4ex}
    \captionof{figure}{
    SSE between the source and target statistics of the FFC-trained (green) and the LFC-trained model (orange). 
    } 
    \vspace{-2ex}
    \Description{Box plots of the sum of squared errors between source and target normalization statistics for each of the 15 CIFAR10-C corruption types, comparing the full-frequency-trained model in green with the low-frequency-trained model in orange. For most corruptions, the orange boxes are lower and narrower than the green ones, meaning the low-frequency-trained model has statistics closer to the source; the largest errors appear for contrast, frost, snow, and brightness.}
    \label{fig:heatmap}
\end{figure}

Figure~\ref{fig:heatmap} further analyzes the interaction between LFC-QAT and FABN. It compares the distance between source BN statistics and target statistics under FFC-trained and LFC-trained models. Across corruptions, the LFC statistics of the LFC-trained model require smaller adaptation than the full statistics of the FFC-trained model. This supports FABN's dual-track update: update LFC statistics conservatively from a source-initialized state, while estimating HFC statistics directly from the current target window.

\begin{table}[t]
    \centering
    \caption{Frequency analysis of FABN on CIFAR10-C showing classification acc. (\%) across different $r$ for low-/high-frequency filters on test data. 
    }
    \vspace{-2ex}
    {\fontsize{9pt}{10pt}\selectfont
    \setlength{\tabcolsep}{1mm}  % 기본값은 6pt 정도
    \begin{tabular}{lcccccccc}
        \toprule
        {} & \multirow{2}{*}{None} & \multirow{2}{*}{FFCs} & \multicolumn{3}{c}{LFCs} & \multicolumn{3}{c}{HFCs} \\
        \cmidrule(lr){4-6} \cmidrule(lr){7-9}
        {} & &  & r=4 & r=8 & r=12 & r=4 & r=8 & r=12 \\
        \midrule
        LSQ & 65.45 &  \textbf{71.22}& 41.94 & 65.11 & 66.59 & 41.78 & 17.41 & 10.66 \\
        LQ & 67.53 & \textbf{76.15} & 42.92 & 70.11& 72.10& 45.38& 16.67& 10.73 \\
        \bottomrule
    \end{tabular}
    }
    \vspace{-2ex}
    \label{tab:results_filters}
\end{table}

%%%%%%%%%%%%%%%%%%%%%%%%%%%%%%
%%%%%%%%%%%%%%%%%%%%%%%%%%%%%%
\subsection{Do LFC-QAT and FABN Each Matter?}

Table~\ref{tab:ablation_components} isolates the contributions of LFC-QAT and FABN. The baseline uses FFC-QAT with NORM. LFC-only replaces FFC-QAT with LFC-QAT while keeping NORM. FABN-only keeps FFC-QAT but replaces NORM with FABN. The full \ours setting uses both LFC-QAT and FABN.
While FABN contributes more than LFC-QAT, both components improve accuracy independently and create synergy when combined together. The results show that \ours's benefit does not come from only one side of the pipeline: source-time LFC-QAT and deployment-time FABN are complementary.

\subsection{Why Split Only the First BN Layer?}

\ours applies frequency decomposition only at the first BN layer. This is a deployment-driven simplification: the first BN input preserves the most direct spatial frequency structure, while applying FABN to every BN layer increases runtime and memory overhead.
Table~\ref{tab:coda*} compares \ours with CoDA$^\ast$, which applies FABN to all BN layers. 
CoDA$^\ast$ achieves nearly the same accuracy as \ours, but substantially increases latency. 
% On Jetson Orin Nano, \ours is about \(1.4\times\)--\(3.0\times\) faster than CoDA$^\ast$ across the reported ResNet18/50 settings, while preserving nearly the same accuracy. 
On Raspberry Pi 4B, \ours is about \(1.6\times\)--\(3.4\times\) faster than CoDA$^\ast$ across the reported ResNet18/50 settings, while preserving nearly the same accuracy. 
% For example, ResNet18 32-bit has 33.87\% accuracy with CoDA$^\ast$ and 33.79\% with \ours, but latency drops from 8.76s to 2.20s. 
% ResNet50 8-bit has 36.65\% accuracy with CoDA$^\ast$ and 36.68\% with \ours, while latency drops from 20.02s to 5.76s. 
% For example, ResNet50 8-bit has 36.65\% accuracy with CoDA$^\ast$ and 36.68\% with \ours, but latency drops from 20.02s to 5.76s. 
For example, ResNet50 8-bit has 36.65\% accuracy with CoDA$^\ast$ and 36.68\% with \ours, but latency drops from 56.04s to 16.58s. 
These results support the first-BN-only design as the better deployment trade-off.

\subsection{What Does Feature Analysis Suggest?}

\begin{table}[t]
    \centering
    \caption{Classification acc. (\%) under varying  ablative settings of \ours with ResNet26 on CIFAR10-C and ResNet50 on ImageNet-C using LQ in 2-bit quantization. 
    }
    \vspace{-2ex}
    \resizebox{\linewidth}{!}{
    \begin{tabular}{l|cc|cc|cc}
        \toprule
        % {}& LFCs & FABN & CIFAR10-C & ImageNet-C & \makecell{ImageNet-C\\(WS=4)} \\
        {}& \multirow{2}{*}{LFCs} & \multirow{2}{*}{FABN} & \multicolumn{2}{c|}{CIFAR10-C} & \multicolumn{2}{c}{ImageNet-C} \\
        & & & $k$=4 & $k$=64 & $k$=4 & $k$=64 \\
        \midrule 
        Baseline (NORM) &  &  & 58.67 & 69.29  & 22.79 & 32.65 \\
        LFC-only & \checkmark &  & 62.92 & 71.94 & 24.00 & 34.51 \\
        FABN-only &  & \checkmark & 72.09 & 74.06 & 30.78 & 35.93 \\
        LFC + FABN & \checkmark & \checkmark & \textbf{74.50} & \textbf{76.15} & \textbf{32.87} & \textbf{38.01} \\
        \bottomrule
    \end{tabular}    
    }
    \label{tab:ablation_components}
    \vspace{-1ex}

\end{table}

\definecolor{gain}{HTML}{1B7837}

\newcommand{\dacc}[1]{\textcolor{black!60}{\scriptsize($\Delta$ #1 pp)}}
\newcommand{\latred}[1]{\textcolor{gain}{\scriptsize($\downarrow$ #1)}}
\newcommand{\latcell}[2]{\cellcolor{gain!8}#1\,\latred{#2}}

\begin{table}[t]
\centering
\caption{
Deployment trade-off between \ours and CoDA$^\ast$.
Values in parentheses on \ours rows are computed relative to CoDA$^\ast$.
(RPi: Raspberry Pi 4B)
}
\vspace{-2ex}
\resizebox{\linewidth}{!}{
\begin{tabular}{llcccc}
\toprule
Model & Method & Acc. (\%) & RPi (s) & Jetson (s) & Laptop (s) \\
\midrule

\multirow{2}{*}{\makecell[l]{ResNet18\\(32-bit)}}
& CoDA$^\ast$ & 33.87 & 18.16 & 8.93 & 2.25 \\
& \ours
& 33.80\,\dacc{-0.08}
& \latcell{11.24}{38.1\%}
& \latcell{6.54}{26.8\%}
& \latcell{1.40}{37.8\%} \\

\midrule

\multirow{2}{*}{\makecell[l]{ResNet18\\(8-bit)}}
& CoDA$^\ast$ & 34.19 & 14.65 & 5.88 & 1.92 \\
& \ours
& 34.17\,\dacc{+0.00}
& \latcell{8.12}{44.6\%}
& \latcell{3.53}{40.0\%}
& \latcell{1.16}{39.6\%} \\

\midrule

\multirow{2}{*}{\makecell[l]{ResNet50\\(32-bit)}}
& CoDA$^\ast$ & 40.76 & 67.29 & 35.16 & 8.46 \\
& \ours
& 40.73\,\dacc{-0.05}
& \latcell{23.96}{64.4\%}
& \latcell{21.33}{39.3\%}
& \latcell{3.83}{54.7\%} \\

\midrule

\multirow{2}{*}{\makecell[l]{ResNet50\\(8-bit)}}
& CoDA$^\ast$ & 36.65 & 56.04 & 21.34 & 7.92 \\
& \ours
& 36.68\,\dacc{+0.03}
& \latcell{16.58}{70.4\%}
& \latcell{7.06}{66.9\%}
& \latcell{3.37}{57.4\%} \\

\bottomrule
\end{tabular}
}
\label{tab:coda*}
\end{table}

\begin{figure}[t]
    \centering
    \includegraphics[width=0.93\linewidth]{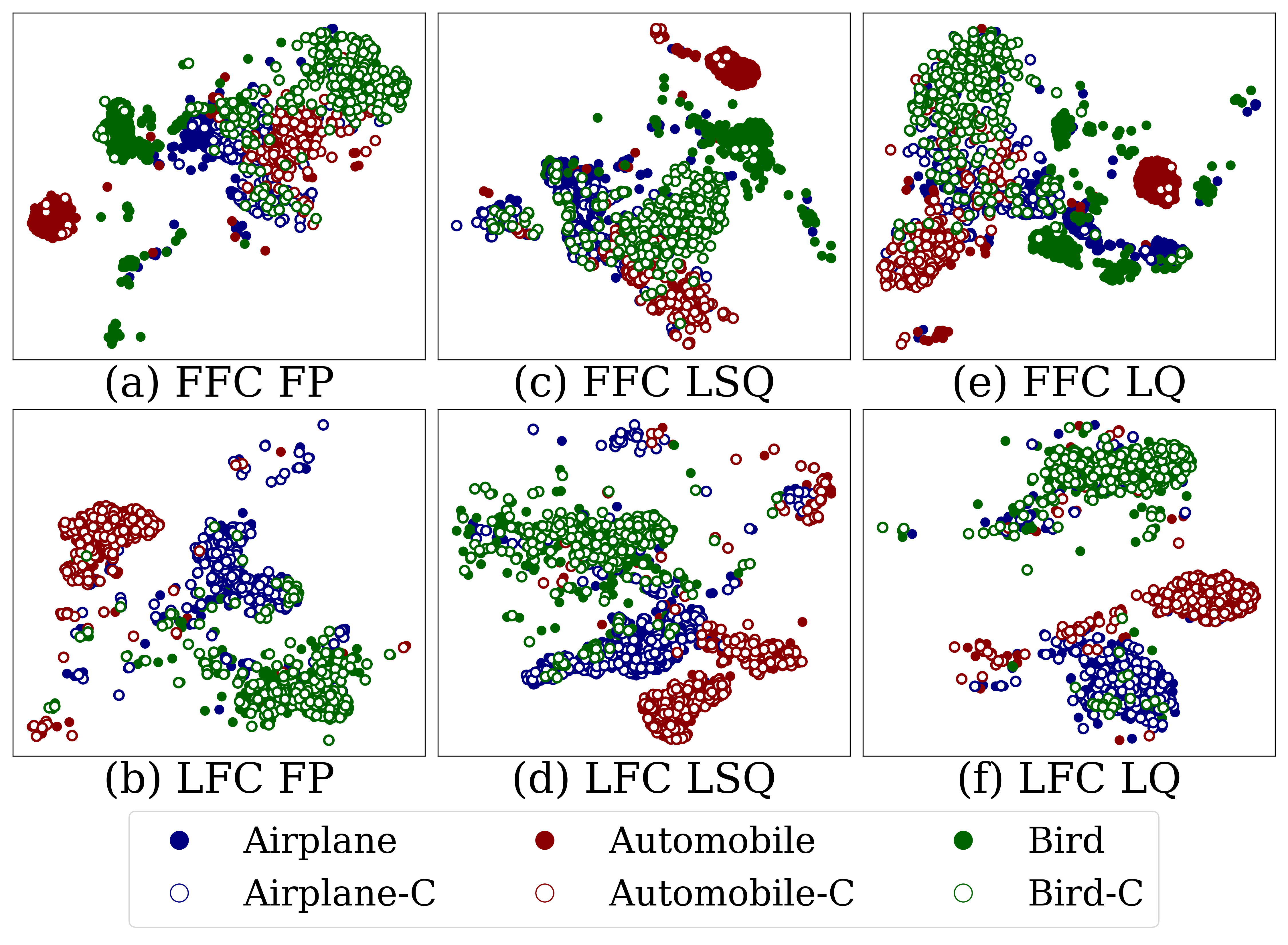} 
    \vspace{-3ex}
    \captionof{figure}{t-SNE visualization of embeddings from FFC/LFC-trained ResNet26 quantized in 2-bit. Tested on three classes (airplane, automobile, and bird) of CIFAR10 (solid) and pixelate in CIFAR10-C (severity 5; hollow). 
    }
    \vspace{-4ex}
    \Description{Six t-SNE scatter plots of embeddings from 2-bit quantized ResNet26, in two rows: full-frequency training on top and low-frequency training below, for FP, LSQ, and LQ. Points are colored by class (airplane in blue, automobile in red, bird in green); filled markers are clean CIFAR10 samples and hollow markers are pixelate-corrupted samples. In the full-frequency plots, clean and corrupted points of the same class often form separate clusters, while in the low-frequency plots clean and corrupted points of each class overlap in one cluster.}
    \label{fig:tsne}
\end{figure}

Finally, we visualize feature embeddings to understand how LFC-QAT affects quantized representations. Figure~\ref{fig:tsne} shows t-SNE~\cite{van2008visualizing} embeddings for three CIFAR10 classes under clean and corrupted inputs. 
FFC-trained QAT models produce more dispersed embeddings under corruption, while LFC-trained models show tighter alignment between clean and corrupted samples within each class. This trend is visible for both LSQ and LQ. The visualization supports the quantitative results above: LFC-QAT makes quantized features less sensitive to corruption shifts, which in turn makes deployment-time statistic adaptation more effective.

%Overall, Section~\ref{sec:analysis} validates the two-stage \ours design. LFC-QAT improves the source-time quantized representation, FABN uses target frequency information to adapt without gradients or weight updates, and first-BN-only FABN preserves the adaptation benefit while reducing deployment overhead.

\section{Related Work}\label{sec:related work}

\noindent\textbf{Efficient and adaptive edge AI.} 
Mobile, embedded, and sensing systems have made DNN inference practical on constrained devices through low-power onboard vision, memory-adaptive execution, dynamic sparsity, mobile GPU scheduling, heterogeneous mobile-processor execution, quantized MCU deployment, and NPU-centric inference~\cite{lee2025hypercam,li2024flexnn,liu2024dynaspa,han2024pantheon,jung2025aria,sun2025adaptqnet,yin2026shadownpu}. Other systems treat post-deployment adaptation as a first-class problem: AdaptiveNet adapts architectures after deployment~\cite{wen2023adaptivenet}, MIRO and DAF support on-device learning/training~{\cite{ma2023cost, liu2025daf}}, AdaShadow accelerates TTA in non-stationary mobile environments~\cite{fang2024adashadow}, and MOCHA adapts video analytics through mobile-cloud collaboration~\cite{zhao2025responsive}. These works optimize execution, scheduling, hardware mapping, or runtime adaptation of deployed models; \ours instead changes what the quantized model learns before deployment so that it can adapt afterward through inference-only state.
%CoDA targets a different point in this design space: it does not perform architecture search, online training, weight updates, expert selection, or cloud collaboration. Instead, it asks whether a single quantized CNN can adapt to domain shift by updating only lightweight inference statistics.

\vspace{1ex}\noindent\textbf{On-device and test-time adaptation.} 
BoTTA recently showed that on-device TTA should be evaluated under realistic edge constraints, including limited target samples, limited category exposure, diverse shifts, overlapping shifts, and real-device resource metrics~\cite{danilowski2026botta}. TinyTTA and AdaShadow further reduce the system cost of adaptation on constrained devices~\cite{jia2024tinytta,fang2024adashadow}. In parallel, TTA methods improve robustness by adapting models with unlabeled target inputs. Backpropagation-based methods such as TENT, EATA, CoTTA, SAR, NOTE, and SoTTA update parameters or auxiliary states using target-domain objectives~\cite{wang2020tent,niu2022efficient,wang2022continual,niu2023towards,gong2022note,gong2023sotta}, but often require gradients, optimizer state, mutable parameters, or full-precision computation. Forward-only methods such as prediction-time BN, covariate-shift adaptation, TEMA, FOA, and backpropagation-free embedding alignment reduce this cost~\cite{nado2020evaluating,schneider2020robustness,su2024unraveling,niu2024forward,ma2026pea}. Recent quantized TTA methods also study adaptation for low-bit models~\cite{deng2025test}. 
\ours is complementary: instead of making post-hoc TTA cheaper for an already trained model, it co-designs the quantized representation and the deployment-time adaptation mechanism.

\vspace{1ex}\noindent\textbf{Quantization-aware training.} 
QAT and integer-only inference reduce model size, memory traffic, and compute cost by training models whose weights and activations can be represented at low precision~\cite{jacob2018quantization,esser2020learned,zhang2018lq,jin2020neural}. Integer-only inference has also been extended to vision transformers~\cite{li2023vit}. These methods are central to edge deployment, but they typically treat deployment as static inference: the model is optimized for source-domain accuracy and then frozen. 
\ours changes what the quantized model learns before deployment. It performs QAT on low-frequency source reconstructions so that the compressed representation emphasizes domain-stable structure and is better aligned with the forward-only adaptation mechanism available after deployment.

\vspace{1ex}\noindent\textbf{Frequency-domain robustness and adaptation.} 
Frequency-domain analysis has shown that spatial frequencies affect robustness and generalization~\cite{yin2019fourier,wang2020high}. Prior work has used Fourier or spectral information for domain adaptation, domain generalization, augmentation, curriculum learning, spectral training, and robustness analysis~\cite{yang2020fda,huang2021fsdr,chen2021amplitude,xu2021fourier,chattopadhyay2023pasta,wang2023efficienttrain,xu2020learning}. 
%
% {While this LFC-stable / HFC-variable premise is revisited in our analysis in Figures ~\ref{fig:distance_matrix} and ~\ref{fig:loss},} 
% \ours uses frequency for a different purpose.
{While this LFC-stable / HFC-variable premise is established by these works rather than new to CoDA, and confirmed in Figures~\ref{fig:distance_matrix} and~\ref{fig:loss}, CoDA uses frequency for a different purpose.}
Rather than using frequency as an augmentation, transformation, curriculum, or source-target alignment tool, \ours uses it to define a deployment-time division of labor: low-frequency source information trains the quantized representation before deployment, while high-frequency target information is reserved for forward-only normalization adaptation after deployment.
%%%%%%%%%% Conclusion
\section{Conclusion}\label{sec:conclusion}

This paper studied adaptive quantized inference for edge vision systems, where models must remain compact for local deployment while adapting to sensing shifts after deployment. \ours addresses this problem through a frequency-asymmetric co-design of source-time training and deployment-time adaptation. Before deployment, LFC-QAT trains the low-bit CNN on low-frequency source reconstructions, encouraging the quantized representation to focus on domain-stable structure. During deployment, FABN updates only normalization statistics from unlabeled target windows, using high-frequency target information to capture current sensing conditions while keeping convolution and linear weights fixed and quantized.

\ours satisfies a strict inference-oriented contract: no source data, no labels, no backpropagation, no optimizer state, no weight updates, and only lightweight mutable inference statistics. Across corruption, rendition, continual, and lightweight-model settings, {and on physically captured lighting shifts,} \ours improves the deployment-compatible accuracy--efficiency trade-off over forward-only baselines, and edge-device measurements show that first-BN-only FABN preserves most of the benefit at low overhead. Adaptive inference thus need not become online training: with the right co-design between offline quantized learning and online statistic adaptation, edge vision models can remain efficient, fixed-weight, and responsive under deployment-time domain shift.

\vspace{1ex}\noindent\textbf{Limitations and discussion.}
{%
\ours is scoped to batch-normalized CNNs, the workload current edge accelerators support~\cite{howard2019mobilenetv3,tan2019efficientnet}} {and also studied in MCU toolchains research~\cite{jacob2018quantization, sun2025adaptqnet, liu2026instmeter}.} {BN statistics are its adaptation interface {— chosen for compatibility with the contract, not as a universally sufficient mechanism}. LFC-QAT is architecture-agnostic {in principle}, but FABN does not {directly} transfer to vision transformers, whose LayerNorm keeps no population statistics and whose token sequences lack the 2D structure the frequency split assumes~\cite{niu2024forward,ma2026pea}.
Synthetic corruptions have frequency signatures that may favor our design. Two results bound this concern: gains shrink on rendition and sketch shifts for \ours and forward-only baselines alike (Table~\ref{tab:results_imagenetrsketch}), yet persist on physically captured lighting shifts (Table~\ref{tab:imagenet-diverse}). \ours thus helps most under sensing-induced shifts and less when a shift alters the learned representation itself {or destroys task-relevant information}.
Class-imbalanced or gradually mixing target streams~\cite{gong2022note,danilowski2026botta} remain future work.%
}

%\noindent\textbf{Limitations and future work.} 
%
%As the first exploration of QAT-TTA co-design, this work focuses on CNNs, widely adopted and studied in both fields. While our study reveals the importance of frequency-aware strategies for the joint design, promising avenues remain open for future investigation. 
%
%Extending \ours to more advanced architectures such as BN-fused CNNs or Vision Transformers is meaningful, yet it will require a different adaptation strategy due to key differences in normalization (BN vs. Fused BN or LayerNorm) and input structures (2D vs. 1D token embeddings).
%%%%%

%%
%% The acknowledgments section is defined using the "acks" environment
%% (and NOT an unnumbered section). This ensures the proper
%% identification of the section in the article metadata, and the
%% consistent spelling of the heading.
\begin{acks}
This work was supported in part by the Institute of Information \& communications Technology Planning \& Evaluation (IITP) grant funded by the Korea government (MSIT) (No. RS-2026-25507282) and the National Research Foundation (NRF) of Korea grant funded by the Korea government (MSIT) (No. RS-2023-00212780, No. RS-2023-00222663).
\end{acks}

%%
%% The next two lines define the bibliography style to be used, and
%% the bibliography file.
\bibliographystyle{ACM-Reference-Format}
% \bibliography{sample-base}
\bibliography{reference}

%%
%% If your work has an appendix, this is the place to put it.
\appendix
\appendix

\section{Baselines} \label{app:reproducibility}
For baselines, we refer to the official implementations of the original authors. 
We use the hyperparameters reported on their paper and code. 
We use torch.optim.SGD for training.
Also, we utilize 2-bit quantized models for LSQ and LQ, unless specified otherwise. 
Additional details of the implementations are provided below.
{The implementation of CoDA is available in the project’s GitHub repository (https://github.com/YoojinKwon/CoDA.git), which contains the source code, experiment configurations, and environment setup instructions. The required software environment setup is listed in requirements.txt. Model architectures and shared utilities are implemented in coda/models/ and coda/modules/, respectively. Training can be launched via coda/train.py or coda/train\_ddp.py for single- and multi-GPU execution, with training scripts  under configs/cifar10/ and configs/imagenet/ specifying the model backbones (ResNet18/26/50, MobileNetV3-small/large, and EfficientNet-B0), quantization settings (FP, LQ, or LSQ), and experiment configurations (ImageNet-C/CIFAR10-C/ImageNet-R/Sketch, ImageNet-ES-Diverse). Adaptation scripts for evaluation are under test/. Reproduction checkpoints (FP baselines, LSQ/LQ models at 2-, 4-, and 8-bit) are hosted on Hugging Face (https://huggingface.co/ideastraw/coda-reproduction-checkpoints).}

\noindent\textbf{FP.} Full-precision models are not quantized.
For CIFAR10, we use learning rate (lr) as 0.1, momentum as 0.9, and weight decay as 1.0e-4. 
The LR scheduler is set to torch.optim.lrscheduler.StepLR with step size 30. 
Batch size is 32 and each model is trained for 90 epochs.
{Data augmentations for Resnet is RandomHorizontalFlip only and for MobileNet and EfficientNet are Resized to 224×224 then RandomHorizontalFlip. Those are repeated in LQ and LSQ.}
For ImageNet, we use lr as 0.1, momentum as 0.9, and weight decay as 1.0e-4. 
LR scheduler is set to torch.optim.lrscheduler.StepLR with step size 30. 
Batch size is 256 and each model is trained for 90 epochs. 
{Data augmentations are RandomResizedCrop to 224×224 then RandomHorizontalFlip for all model types. Those are repeated in LQ and LSQ.}

\noindent\textbf{LSQ.} Following the original paper,~\cite{esser2020learned}, we initialize step size $s$ as $s$=$2\overline{w}/\sqrt{p}$ where $\overline{w}$ refers to the mean of weights in the corresponding layer and $p$ is $2^{\text{bitwidth}-1}-1$. 
We use the reported hyperparameter setting which showed the best accuracy by each hyperparameter. 
For CIFAR10, we use learning rate (lr) as 0.1, momentum as 0.9, and weight decay as 5.0e-4. 
LR scheduler is set to MultiStepLR of torch.optim.lrscheduler with milestones of [10, 30, 50, 70]. 
Batch size is 256 and each model is trained for 90 epochs. 
For ImageNet, we use learning rate (lr) as 0.01, momentum as 0.9, and weight decay as 2.5e-4. 
LR scheduler is set to CosineAnnealingLR of torch.optim.lrscheduler with Tmax is 90. 
Batch size is 256 and each model is trained for 90 epochs. {For all evaluated architectures, the first and last layers are kept at 8-bit precision for both weights and activations, following the original LSQ implementation.}

\noindent\textbf{LQ.} For CIFAR10, we use learning rate (lr) as 0.1, momentum as 0.9, and weight decay as 1.0e-4. 
LR scheduler is set to MultiStepLR of torch.optim.lrscheduler with milestones of [82, 123]. 
Batch size is 128 and each model is trained for 200 epochs. 
For ImageNet, we use learning rate (lr) as 0.01, momentum as 0.9, and weight decay as 1.0e-4. 
LR scheduler is set to MultiStepLR of torch.optim.lrscheduler with milestones of [10, 30, 60, 80, 95, 105]. 
Batch size is 256 and each model is trained for 120 epochs. {For all evaluated architectures, the first and last layers are kept in full precision for both weights and activations, following the original LQ implementation.}

\section{Inter-domain Distance Calculation}\label{sup:distance_calc}

To compare how domain-invariant the LFC and HFC are, we calculate the inter-domain distances for each and constructed a matrix based on these values.
From the CIFAR10-C dataset, we select one class and randomly choose a single clean original image from this class, along with its corresponding 15 corrupted versions. 
Each corrupted image is transformed into frequency domain features using FFT. 
We then apply LPF and HPF to these frequency domain features, extracting LFC and HFC in frequency domain, respectively.
To quantify the variation of LFC and HFC features across different domains, we calculate the cosine distances separately between the LFC features and between the HFC features of images with different corruptions. 
This results in a $15 \times 15$ matrix for each feature type. 
We repeat this process for 10 randomly selected samples within the same class, averaging their resulting matrices. 
Finally, we perform this procedure for all classes, obtaining the overall inter-domain cosine distance matrix.
A distance matrix with larger values indicates that the corresponding frequency component exhibits greater variability across domains.
As in Figure~\ref{fig:distance_matrix}, the values in the LFC distance matrix are relatively small, indicating that the LFC is more domain-invariant compared to the HFC.

\section{License of Assets} \label{supp: license}

\noindent\textbf{Datasets} CIFAR10 (MIT License), CIFAR10-C (Apache 2.0), ImageNet-C (Apache 2.0), ImageNet-R (MIT License), ImageNet-Sketch (MIT License){, ImageNet-ES-Diverse (CC BY 4.0).}

\noindent\textbf{Codes} Code for Fourier transformation and low-/high-pass filtering of EfficientTrain (MIT License), 
torchvision for ResNet18 and ResNet50 (Apache 2.0), 
official repository of LQ (MIT License), 
official repository of TEMA (MIT License),
official repository of TENT (MIT License), 
official repository of SAR (BSD 3-Clause License), 
and the official repository of CoTTA (MIT License).

\end{document}